\documentclass[10pt,twocolumn,letterpaper]{article}

\usepackage{cvpr}      

\usepackage[dvipsnames]{xcolor}

\usepackage{amsmath}
\usepackage{amssymb}
\usepackage{bm}
\usepackage{multirow}
\usepackage[ruled,vlined]{algorithm2e}
\usepackage{algpseudocode}
\usepackage{subcaption}

\usepackage{booktabs}
\usepackage{xspace}
\usepackage{url}

\usepackage{graphicx}
\usepackage{lipsum}
\usepackage{microtype}
\usepackage{nicefrac}
\usepackage{verbatim}
\usepackage{bbding}

\usepackage{color, soul}
\usepackage{listings}

\newcommand{\describeContent}[1]{%
\begingroup%
\let\thefootnote\relax%
\footnotetext{#1}%
\endgroup%
}

\definecolor{cvprblue}{rgb}{0.21,0.49,0.74}

\usepackage[pagebackref,breaklinks=true,colorlinks,citecolor=cvprblue,bookmarks=false]{hyperref}
\usepackage{listings}
\usepackage[accsupp]{axessibility} 
\def\paperID{10430} 
\def\confName{CVPR}
\def\confYear{2024}

\title{MULAN: A Multi Layer Annotated Dataset for Controllable Text-to-Image Generation}
\author{Petru-Daniel Tudosiu*\textsuperscript{1} \qquad Yongxin Yang\textsuperscript{1} \qquad Shifeng Zhang\textsuperscript{1} \qquad Fei Chen\textsuperscript{1} \\ \qquad Steven McDonagh\textsuperscript{2 $\dag$}   \qquad Gerasimos Lampouras\textsuperscript{1} \qquad Ignacio Iacobacci\textsuperscript{1} \qquad Sarah Parisot*\textsuperscript{1} \\ \\ 
\textsuperscript{1}Huawei Noah's Ark Lab, 
\textsuperscript{2}University of Edinburgh}

\begin{document}
\maketitle
\describeContent{* equal contribution, $\dag$ work done in part at Huawei Noah's Ark Lab}

\begin{abstract}
Text-to-image generation has achieved astonishing results, yet precise spatial controllability and prompt fidelity remain highly challenging. This limitation is typically addressed through cumbersome prompt engineering,  scene layout conditioning, or image editing techniques which often require hand drawn masks. Nonetheless, pre-existing works struggle to take advantage of the natural instance-level compositionality of scenes due to the typically flat nature of rasterized RGB output images. Towards adressing this challenge, we introduce MuLAn: a novel dataset comprising over 44K MUlti-Layer ANnotations of RGB images as multi-layer, instance-wise RGBA decompositions, and over 100K instance images. To build MuLAn, we developed a training free pipeline which decomposes a monocular RGB image into a stack of RGBA layers comprising of background and isolated instances. We achieve this through the use of pretrained general-purpose models, and by developing three modules: image \emph{decomposition} for instance discovery and extraction, instance \emph{completion} to reconstruct occluded areas, and image \emph{re-assembly}. We use our pipeline to create MuLAn-COCO and MuLAn-LAION datasets, which contain a variety of image decompositions in terms of style, composition and complexity. With MuLAn, we provide the first photorealistic resource providing instance decomposition and occlusion information for high quality images, opening up new avenues for text-to-image generative AI research. With this, we aim to encourage the development of novel generation and editing technology, in particular layer-wise solutions. MuLAn data resources are available at \url{https://MuLAn-dataset.github.io/}

\end{abstract}    
\section{Introduction}
\label{sec:intro}

\begin{figure}
    \centering
    \begin{subfigure}[t]{\linewidth}
    \centering\includegraphics[width=0.95\linewidth]{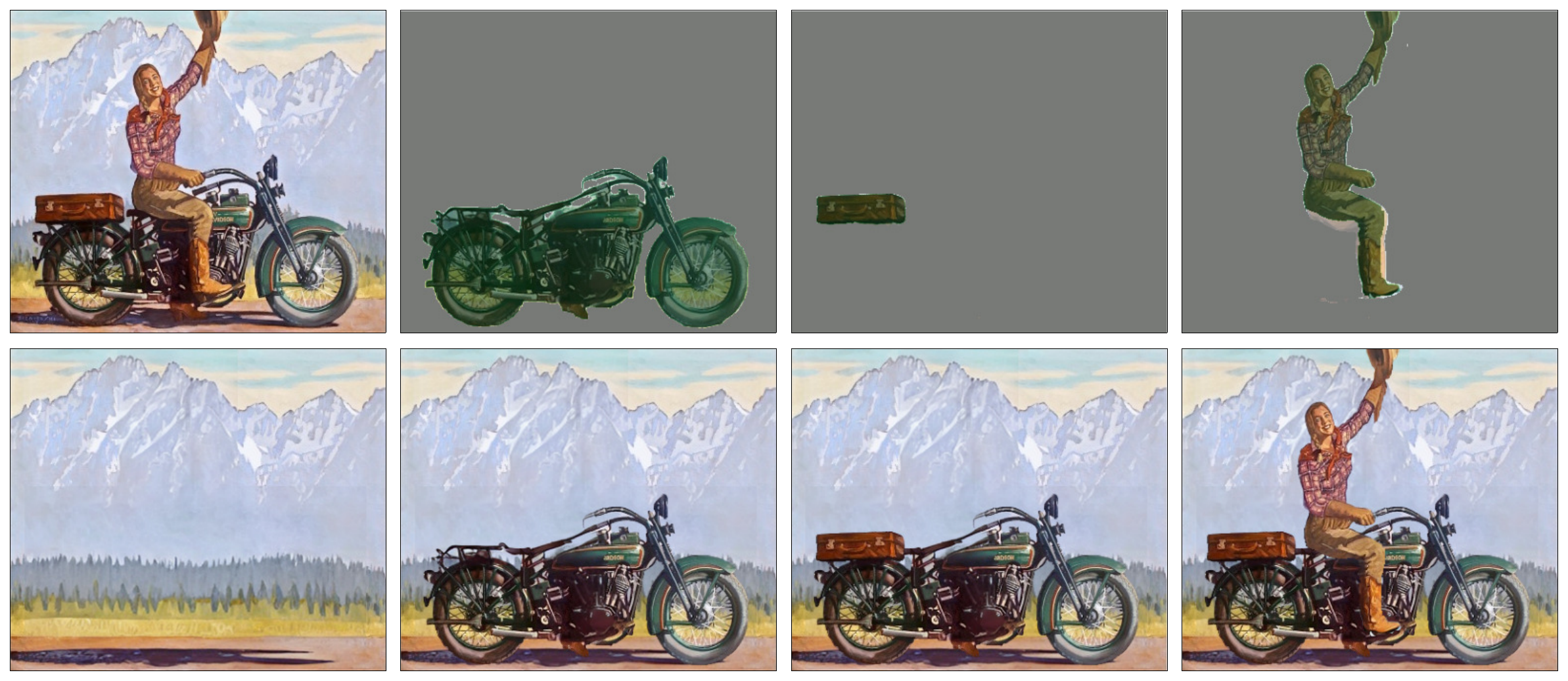}
    \end{subfigure}
    \begin{subfigure}[t]{\linewidth}
    \centering\includegraphics[width=0.95\linewidth]{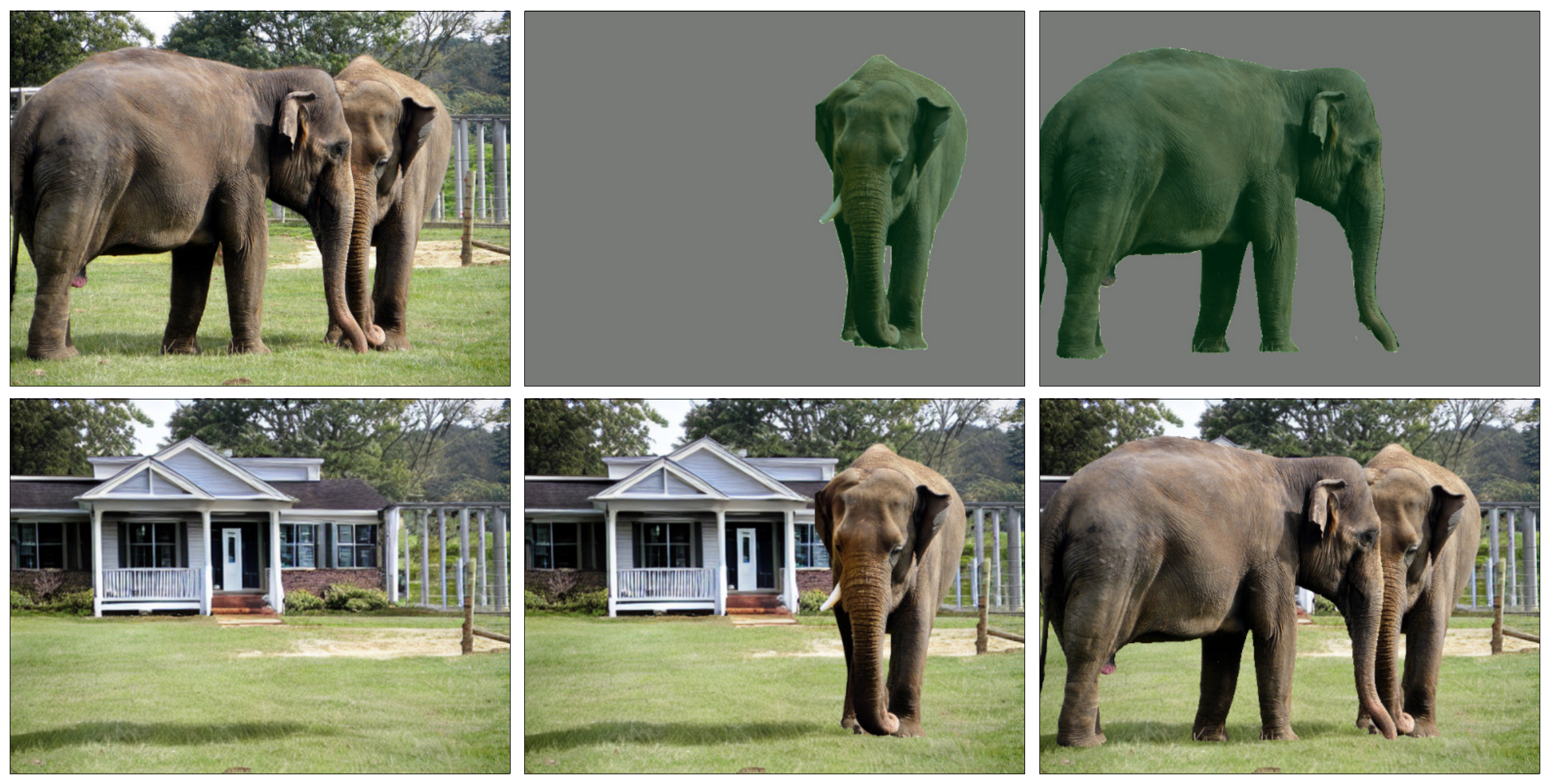}
    \end{subfigure}
    \caption{Example annotations from our MuLAn dataset. We decompose an image into a multi-layer RGBA stack, where each layer comprises an instance image with transparent alpha layer (green overlays) and background image. For each scene, the second row shows iterative addition of RGBA instance layers.}
    \label{fig:teaser}
\end{figure}

\looseness=-1
Large scale generative diffusion models \cite{rombach2022high, podell2023sdxl} now enable generation of high quality images from text prompt descriptions. These models are typically trained on large datasets of captioned RGB images encompassing multiple styles and contents. While such techniques have pushed the field of text-guided image generation forward tremendously, precise controllability of image appearance and composition (\eg local image attributes, countability) still remains a challenge \cite{huang2023t2icompbench}. Prompt instructions can often lack precision or be misunderstood (\eg counting errors, incorrect spatial positions, bleeding of concepts, failure to add or remove instances), and therefore require intricate prompt engineering to obtain the desired result. Fine-tuning a generated image by 
even slightly changing the prompt can result in a 
markedly different sample, further increasing the amount of effort required to obtain the desired image. 

Efforts towards addressing these limitations have considered additional conditioning in the form of \eg~poses, segmentation maps, edge maps~\cite{zhang2023adding, mou2023t2i} and model-based image editing strategies~\cite{couairon2022diffedit, hertz2022prompt, epstein2023diffusion}. The former improves spatial controllability, yet still requires tedious prompt engineering to adjust image content; while the latter often fails to understand spatial instructions and therefore struggles to accurately modify desired image regions without affecting other areas or introducing unwanted morphological changes.

We conjecture that a key obstacle is the typically flat 
nature of rasterised RGB images, which fails to leverage of the compositional nature of scene content. Alternatively, isolating instances and background on individual RGBA layers has the potential to grant precise control over image composition, as processing of instances on separate layers guarantees
content preservation. This can trivialise image manipulation tasks like resizing, moving, or adding/removing elements, which remain a challenge for current editing methods.
 
Collage Diffusion~\cite{sarukkai2023collage} and Text2Layer~\cite{zhang2023text2layer} have shown preliminary evidence of the benefits of multi-layer composable image generation.  Collage diffusion controls image layout by composing arbitrary input layers \eg by sampling composable 
foreground and background layers, while Text2Layer explores decomposition of images into two separate layers (grouping foreground instances and background). Despite an increasing interest in training-free layered and composite generation \cite{tficon, layerdiffusion}, a major barrier to research development in this promising direction is the lack of publicly available photorealistic, multi-layer data to train and evaluate generative and editing methodology. 

\begin{figure*}[t]
\centering
\begin{minipage}[b]{.23\textwidth}
\centering
\includegraphics[width=0.80\linewidth]{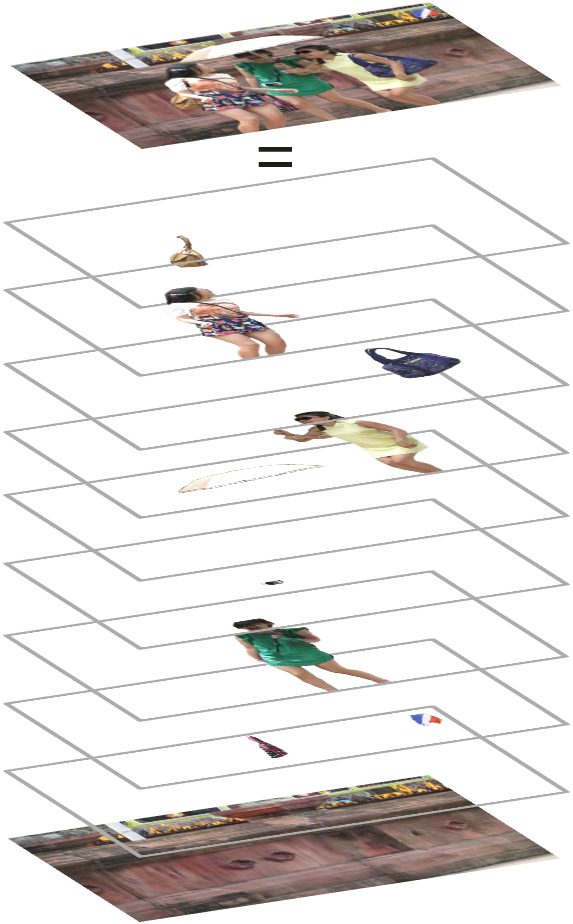}
\caption{Illustration of our RGBA decomposition objective.}
\label{fig:decomp}
\end{minipage}
\hfill
\begin{minipage}[b]{.75\textwidth}
\includegraphics[width=\linewidth]{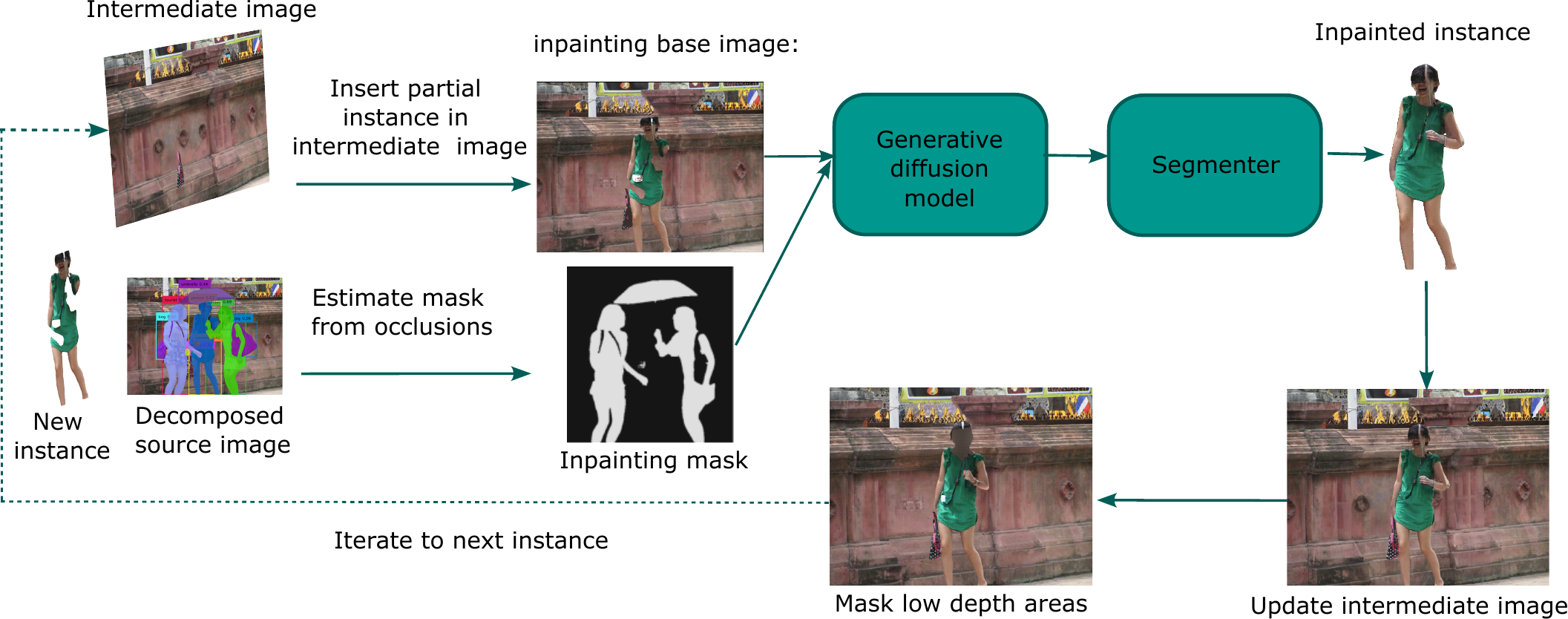}
\caption{Illustration of the inpainting procedure for a given instance. }
\label{fig:inpaint}
\end{minipage}
\end{figure*}

\looseness=-1
In this work, we aim to fill this gap by introducing MuLAn, a novel dataset comprising of 
multi-Layer RGBA 
decomposition annotations of natural 
images (see Fig.~\ref{fig:decomp} for an RGBA decomposition illustration). To achieve this, we design an image processing pipeline that takes as input a single RGB image and outputs a multi-layer RGBA decomposition of its background and individual object instances.  
We propose to leverage large-scale pre-trained foundational models to build a robust, general purpose pipeline 
without incurring any additional model training costs. We separate our decomposition process into three submodules, focusing on 1) instance discovery, ordering and extraction, 2) instance completion of occluded appearance, and 3) image re-assembly as an RGBA stack. Each submodule is carefully designed to ensure general applicability, high instance and background  reconstruction quality, and maximal consistency between input image and composed RGBA stack. 
We process images from the COCO~\cite{coco} and LAION Aesthetics 6.5~\cite{schuhmann2022laion} datasets using our novel pipeline, yielding multi-layer instance annotations for over 44K images and over 100K instances. Illustrations of generated decompositions are shown in Fig.~\ref{fig:teaser}: each decomposed image comprises a background layer, and extracted instances are separate RGBA images with transparency alpha layers. Instances can be removed from the RGBA stack, yielding several intermediate representations; 
where resulting occluded areas 
are completed via inpainting. 

Our goal in releasing MuLAn, is to foster development and training of technologies to generate images as RGBA stacks, by offering comprehensive scene decomposition information and scene instance consistency. We aim to facilitate research seeking to ($i$) advance controllability of generated image structures 
and ($ii$) improve local image modification quality, via precise layer-wise instance editing. 
This paper illustrates the potential utility of our dataset and the benefits of layer-wise representations through two applications: 1) RGBA image generation and 2) instance addition image editing. In summary, our main contributions are: 
\begin{itemize}
    \item The release of MuLAn, a novel dataset of multi-layer annotations, comprising RGBA decompositions of over 44K images, derived from COCO and LAION Aesthetics 6.5. To the best of our knowledge, MuLAn is the first dataset of its kind, providing instance decomposition and occlusion information for a large variety of scenes, styles (including photorealistic real images), resolutions and object types. 
    \item A novel, modular pipeline that decomposes single RGB images into instance-wise RGBA stacks at no additional training cost. Our idea leverages large pre-trained models in an innovative manner, and comprises ordering and iterative inpainting strategies to achieve our image decomposition objective. This further enables unique insight into the behaviour of popular large-models in the wild.
    \item We showcase MuLAn's potential through two applications 
    that leverage our rich annotations in distinct ways.
\end{itemize}

\begin{table}[t]
\centering
\resizebox{\linewidth}{!}{%
\begin{tabular}{l|cccccccc}
\toprule
Dataset                       & \# Images & Resolutions & \# Classes          & \# Instances & \begin{tabular}[c]{@{}c@{}}Occluded\\ Instances\end{tabular} & \begin{tabular}[c]{@{}c@{}}Average\\ Occlusion Rate\end{tabular} & Synthetic & Ordering  \\ \midrule
SAIL-VOS \cite{hu2019sail}    & 111,654   & 800x1280    & 162                 & 1,896,296    & 1,653,980                                                    & 56.3 \%                                                          & $\surd$   & $\surd$   \\
OVD \cite{yan2019visualizing} & 34,100    & 500x375     & 196 (vehicles)     & -            & -                                                            & -                                                                 & $\surd$   & -         \\
WALT \cite{reddy2022walt}     & 15 Mil    & 4K/1080p    & 2 (vehicles)        & 36 Mil       & -                                                            & -                                                                & partially & $\surd$   \\
AHP \cite{zhou2021human}      & 56,599    & Non-fixed   & 1  (humans)         & 56,599       & -                                                            & -                                                                & partially & -         \\  
DYCE \cite{ehsani2018segan}   & 5,500     & 1000x1000   & 79 (indoor scenes)  & 85,975       & 70,766                                                       & 27,7\%                                                           & $\surd$   & -         \\
OLMD \cite{dhamo2019object}   & 13,000    & 384x512     & 40 (indoor scenes)  & -            & -                                                            & -                                                                & $\surd$   & $\surd$   \\
CSD \cite{zheng2021visiting}  & 11,434    & 512x512     & 40  (indoor scenes) & 129,336      & 74,596                                                       & 26.3\%                                                           & $\surd$   & $\surd$   \\ \midrule
MuLAn-COCO                    & 16,034    & Non-fixed   & 662                 & 40,335       & 15,223                                                       & 7.2 \%                                                           & partially & $\surd$   \\
MuLAn-LAION                   & 28,826    & Non-fixed   & 705                 & 60,934       & 14,009                                                       & 8.2 \%                                                           & partially & $\surd$   \\
MuLAn                         & 44,860    & Non-fixed   & 759                 & 101,269      & 29,232                                                       & 7.7 \%                                                           & partially & $\surd$
\end{tabular}%
}
\caption{Comparison between MuLAn and publicly available related datasets providing amodal masks and appearance information. }
\label{tab:datasets_stats}
\end{table}

\section{Related Work} \label{sec:rel}

\noindent\textbf{Amodal completion} aims to automatically estimate the real structure and appearance of partially occluded objects. This challenging task has been heavily researched \cite{ao2023image}, typically building models that are trained on synthetic or richly annotated datasets. Such datasets typically comprise instance segmentation masks that include occluded regions. In addition, the closest datasets to MuLAn comprise appearance information of occluded areas and instance ordering information. We provide a detailed comparison of these datasets with ours in Tab.~\ref{tab:datasets_stats}. Time and cost requirements of producing ground truth amodal annotations have limited previous research to synthetic, small or highly specialised datasets such as indoor scenes~\cite{ehsani2018segan, zheng2021visiting, dhamo2019object}, humans~\cite{zhou2021human}, vehicles \cite{yan2019visualizing}, and objects and humans \cite{reddy2022walt,hu2019sail}. In contrast, MuLAn comprises images with a large variety of scenes, styles (including photorealistic real images), resolutions and object types; and was built on top of popular datasets to support generative AI research. We highlight that our use of real images impacts occlusion rate compared to existing datasets, where synthetic scenes were purposely designed to have a high rate.

\looseness=-1
\noindent\textbf{RGBA image decomposition} requires identifying and isolating image instances on individual transparent layers, and estimating the shape and appearance of occluded areas. This challenging task is typically carried out using additional inputs (beyond a single RGB image), such as inmodal segmentations ~\cite{zhan2020deocclusion}, stereo images ~\cite{gonzalez2022sainet} and temporal video frames. The latter substantially facilitates the decomposition tasks, as video frames can provide missing occluded information~\cite{lu2020layered, tukra2021see}.
Recently, layer based generative modelling benefits from initial explorations. Text2Layer~\cite{zhang2023text2layer} 
creates a two-layer RGBA decomposition of natural images. Images are decomposed into a background and salient foreground layer, where the background is inpainted using prompt-free state-of-the-art diffusion models. The main limitation of this, compared to our approach, is 
the two-layer decomposition: all instances are extracted in the same foreground layer which critically lacks our required flexibility of instance wise decomposition.
Our objective, to decompose each instance individually, comes with additional challenges such as instance ordering, instance inpainting and amodal completion. Adjacent to our decomposition objective, PCNet~\cite{zhan2020deocclusion} learns to predict instance ordering, amodal masks and object completion. The approach's applicability is however restricted by the aforementioned limitations of amodal completion datasets. To the best of our knowledge, our decomposition pipeline is the only general purpose technology capable of decomposing monocular RGB images.

\looseness=-1
Complementary to our work, Collage Diffusion~\cite{sarukkai2023collage}, an image collage strategy for diffusion generative models, is developed with a similar instance-level modularity objective. 
While we aim to extract instances from an image, their method seeks to assemble individual instances in a homogeneous composite image. One limitation of this prior work involves the challenge of balancing appearance preservation of collaged instances with homogeneity of the composite image, which can be considered non-trivial and increases in difficulty with instance count. 
\section{Image Decomposition Pipeline}\label{sec:method}

Our objective is to decompose a single RGB image $I$ into an instance-wise stack of $N$ RGB-A image layers \mbox{$\mathcal{S}=\{l_i \,|\, i \in 1,\dots, N\}$}, where the A-layer (Alpha) describes the transparency of each RGB instance. As illustrated in Fig. \ref{fig:decomp}, each layer $l_i$ comprises either a background image, or a single instance with full transparency in non-instance regions. Flattening $S$ should yield our original image $I$. Due to the lack of large, general purpose datasets to provide this level of granular information, we approached this objective in a training free manner, leveraging a combination of specialised large scale pre-trained models.

\begin{figure}
    \centering
    \includegraphics[width=0.6\linewidth]{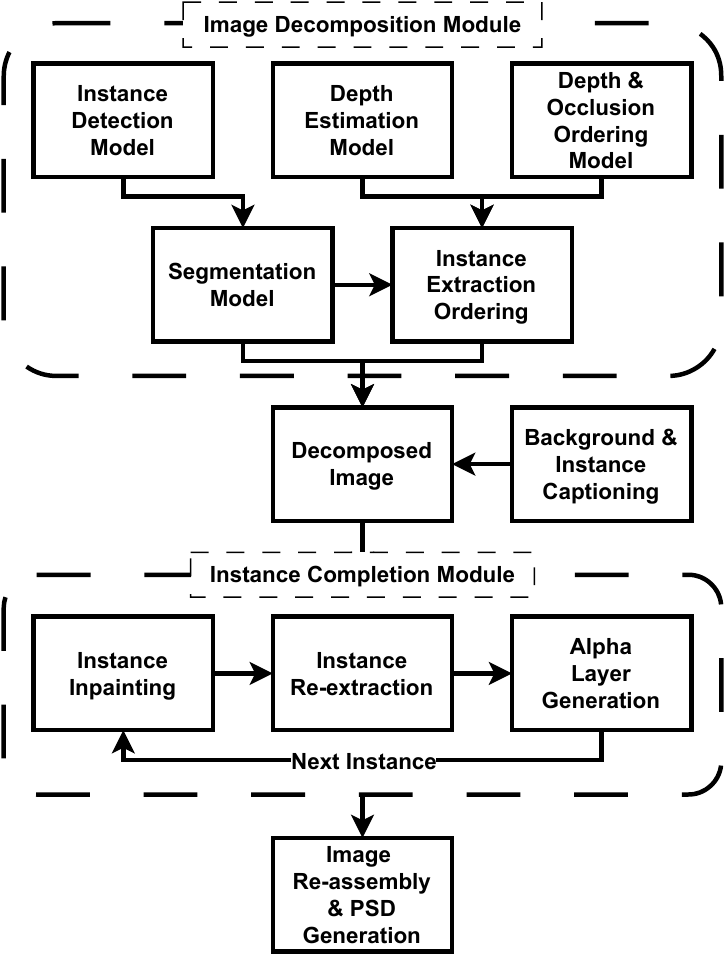}
    \caption{Overview of our RGBA decomposition pipeline}
    \label{fig:pipeline}
\end{figure}

\looseness=-1
We build a pipeline that comprises a sequence of three main modules. First, the \emph{Decomposition} module encompasses instance discovery and extraction. It focuses on scene understanding, comprising a sequence of object detection, segmentation and depth estimation models. This module decomposes $I$ into a background image and a series of extracted instances with associated segmentation masks. At this stage, background and extracted instances are missing information due to occlusions. This is addressed with our second main module: \emph{Instance Completion}. 
This second component aims to reconstruct each instance individually, as it would look like if there were no occlusions. This step leverages depth and relative occlusion information to establish an instance ordering, and state-of-the-art text-to-image generative models to inpaint occluded areas.
Finally, the image \emph{Reassembly} module generates occlusion aware Alpha layers, and builds our RGBA stack such that flattening the stack effectively reconstructs $I$. 

An overview of our pipeline is found in Fig.~\ref{fig:pipeline} and a further detailed schematic, showing all components' instantiations, is also available in the supplementary materials. 

\subsection{Image Decomposition Module}

Our decomposition module aims to extract and isolate all instances in the image. We first identify and segment instances using object detection and segmentation models. In parallel, we rely on depth estimation and occlusion ordering models to build relative occlusion graphs, and establish an instance ordering for extraction, inpainting and reassembly. 

\noindent \textbf{Object Detection.} Accurately detecting all relevant instances in an image is the first step of our pipeline. In order to achieve good quality decomposition, it is essential that we are able to detect and separate all instances present in the scene. To this end, we leverage vision-language object detection techniques, that input a list of categories to detect, alongside the input image. Such models are attractive as they easily allow open-set detection, meaning we are not limited to the pre-existing class sets of specific data. 
We use detCLIPv2 \cite{yao2022detclip, yao2023detclipv2}, a state-of-the-art model characterised by its ability to leverage category definitions (and not just class names) to improve detection accuracy. We carefully construct our text input (the category list), to ensure all desired categories are detected and extracted from the image. We use the concept list from the THINGS~\cite{hebart2019things} database, and manually update and simplify it to obtain more generic category names (\eg merging types of boats, drinks, nuts, etc.), and remove homonyms and concepts that we do not want to extract (\eg unmovable objects, clothing, bolts and hinges). 
We highlight that this list constitutes an input to the pipeline, easily allowing customisation of which instances to detect. This class list is used, alongside definitions from the WordNet~\cite{miller1995wordnet} database, to identify all relevant instances in an image. This step of the pipeline outputs a series of bounding boxes with corresponding category names.

\looseness=-1
\noindent \textbf{Segmentation.} Our next step is to precisely segment detected instances. In order to handle a large number of categories, domains and image qualities we seek to leverage a robust general purpose segmentation model. 
One such model is SAM \cite{kirillov2023segment}, which has been trained with the required diversity and scale, achieving good robustness and transferability across a large set of domains. 
The ability to use bounding boxes as grounding for segmentation predictions makes this family of models an excellent candidate to be combined with our detCLIPv2 detector.

\looseness=-1
\noindent \textbf{Depth Estimation.} 
Understanding the relative positioning of instances in an image is crucial towards achieving our RGBA decomposition goal. Depth estimation provides essential information, indicating distance to the camera at capture time.  
We use the MiDaS model~\cite{Ranftl2022}, 
chosen for its robustness: it was trained on 12 different datasets allowing it to be reliable across different type of scenes and image qualities. Once computed, we quantise the depth map into multiple bins of width 250 relative depth units to facilitate cross instance comparisons. 

\looseness=-1
\noindent \textbf{Instance Extraction.} We define instance extraction as the application of a binary mask onto the full image in order to isolate a detected instance from the rest of the image. We employ a set of strategies to increase the robustness of this crucial step. First, we estimate a proto-ordering by clustering instances based on their bounding box overlap, and use bounding box size and mean depth value (within the segmentation mask) to order them. 
Second, we use our proto-ordering 
to enforce disjoint instance segmentation masks by excluding extracted areas from the masks of following instances. 
Lastly, 
instances whose largest connected component is smaller than 20 pixels or 0.1\% of the entire image are not extracted. 

\noindent \textbf{Depth and Occlusion Graphs.} We further compute depth and occlusion instance graphs for a more comprehensive scene understanding. Specifically we are using the $\text{InstaDepthNet}^{o,d}$ model \cite{lee2022instance}, which is 
capable of jointly predicting occlusion and depth relationships between instances. The model predicts instance pairwise relationships, using the original image and instance segmentations as input. The directed occlusion graph outlines relative occlusion information between instances. Image instances are represented by graph nodes, and a directed edge from instance A to instance B (A $\rightarrow$ B) indicates that A occludes B. We note that valid bidirectional edges exist where two instances mutually occlude one another (A $\leftrightarrow$ B). 
Similarly, the directed depth graph also represents instances as nodes, and A $\rightarrow$ B indicates that A is closer to the camera than B, as defined by instance mean depth values. A bidirectional edge (A $\leftrightarrow$ B) indicates that both instances have the same mean depth.

\noindent \textbf{Instance ordering.} In order to maximise instance completion quality, inpainting of occluded areas is performed using contextual information from the original image. 
As a result, establishing a precise instance inpainting schedule is crucial towards progressively enriching the image context without occluding relevant areas. We generate our instance ordering in three steps, relying on depth ordering and occlusion information obtained in our decomposition step. First, instances are ordered based on their depth information, from furthest away to closest (according to instance mean depth value). This can easily be achieved using the instance depth graph, by computing node out-degree: this computes the number of directed edges departing a node, \ie~the number of instances that are behind our node. Second, we rely our occlusion graph to refine our ordering: if instance A occludes instance B, instance B will systematically be ordered before instance A. Finally, mutually occluded instances are reordered according to their maximum depth value. Instance ordering algorithm details are provided in Supplementary materials.

\subsection{Instance Completion Module}

Prior to instance completion, we have successfully detected, isolated and ordered all instances from the background image. 
An important challenge remains: reconstructing occluded areas for each image layer $l_i$  individually (including the background), such that removing or hiding any layer reveals occluded areas. Since we are decomposing natural images, this information is not available to us. We rely on state-of-the-art generative models to imagine these occluded areas from available context, using inpainting.  

\looseness=-1
Diffusion model based image inpainting has set a new standard in comparison with traditional inpainting techniques~\cite{buyssens2015exemplar,jin2015annihilating} as they take advantage not only of image contents but also of a learned image prior and textual conditioning. Even so, our setting comes with unique difficulties: 1) in contrast with the common strategy of carefully engineering hand-crafted prompts, we can only rely on automatically generated captions, 
2) instance images comprise the instance on a background of uniform colour, an image mode not commonly seen by these models, and 3) rather than obtaining beautiful or creative images, we seek the simple, accurate and high quality completion for our available content. We next provide detail on our inpainting process and how these difficulties are addressed. 

\noindent \textbf{Inpainting procedure.} An overview of our inpainting process is illustrated in Fig.~\ref{fig:inpaint}. Given a pre-defined instance ordering, we iteratively inpaint an instance's occluded areas, starting from the background image to the closest 
instance. For a given instance, our inpainting process proceeds as follows: we first estimate an inpainting mask using occlusion ordering information and segmentation masks from occluding instances. Second, we build a contextual inpainting image by re-integrating our incomplete instance in an intermediate background image. This background image contains inpainted instances processed in previous iterations. Third, the instance is inpainted using a state-of-the-art inpainting generative model and automatic captions as prompts. Fourth, we re-extract the completed instance using our segmentation model and the occluded segmentation mask as guidance, effectively obtaining the complete instance image which will be part of our multi-layer representation. Finally, we update our background inpainting image for the next iteration by integrating our newly inpainted instance. 
Importantly, we aim to strike a balance between maximising scene context and preventing introduction of irrelevant image content. 
This is particularly important for mutually occluded instances: for example, considering a person holding a phone in their hands, with the person's hand as context, fingers will be reconstructed when inpainting the phone's occluded areas. To prevent this, we ``hide'' potentially misleading context by replacing information from pixels that have higher depth than the next instance's max depth with a constant value.

\noindent \textbf{Inpainting mask.} Estimating an accurate inpainting mask, \ie~describing which image regions will be overwritten, is crucial towards achieving accurate instance completion. Failing to include key occluded areas risks yielding incomplete results, while a mask that is too large risks altering the original image appearance. Ideally, one would estimate an accurate complete instance shape via amodal completion techniques~\cite{ao2023image}. Existing methods tend, however, to be dataset or object class specific with limited generalisation ability. We alternatively propose to leverage instrinsic biases of large generative models by providing a large inpainting mask, encompassing the area where the occluded object \emph{could} be present. This is achieved by building an inpainting mask comprising segmentation masks of all occluding instances.  

\noindent \textbf{Inpainting prompts} remain simple, as we seek a fully automated decomposition strategy. 
For instance inpainting, we leverage automatically generated instance captions (see Sec.~\ref{sec:caption}). 
To inpaint the background image, we use a simple generic prompt (\textit{``an empty scene''}) that ensures the generated inpainting background is as simple as possible. Importantly, we include class names of all other instances in all negative prompts, to avoid re-introducing extracted instances. This increases robustness to imperfect segmentations.    

\subsection{Image re-assembly module}

The last and simplest module re-assembles all individual RGB images into an ordered RGBA stack that, once flattened, yields an image as close as possible to the the original input image. Instance RGB images are ordered following our inpainting ordering, such that instances inpainted last are at the top, with the background at the bottom of the stack. Following this order, we iteratively generate Alpha layers for each stack element by refining instance segmentation masks. 
We 
post process SAM segmentations obtained after inpainting with the image matting model VitMatte~\cite{yao2023matte} to improve alpha blending quality, handle transparent objects, and address SAM undersegmentation tendencies. 
While undersegmenting is preferred for the first two modules, in order to avoid introducing proximal content and erroneous priors when inpainting, we require accurate delineations for this last stage. VitMatte refines SAM outputs, providing smoother non-binary segmentations, and allows us to blend the inpainted instances in a more natural way. 
In settings where mutual occlusion exist (\ie~a lower level instance is creating an occlusion), we further adjust alpha layers by setting occluded areas as transparent. 
This last module finally outputs our RGBA stack image decomposition.

\subsection{Captioning strategy} 
\label{sec:caption}

We generate captions for all layers (background, instance), intermediate flattened RGBA stacks and the full image. We use LLaVa \cite{liu2023improved} to generate detailed captions for standard images. Due to the unique nature of instance images (an instance on a uniform white background), verbose captioning models like LLaVa tend to hallucinate image features. To address this, we leverage a BLIP-2 model \cite{li2023blip} to caption instances and performed a grid search to select a parameter set limiting verbosity and hallucination. Furthermore, we use constricted beam search to generate multiple captions and select the best one with CLIP~\cite{radford2021learning}. Components captioned with LLaVa are also captioned with BLIP, for completeness.

\section{MuLAn Dataset}\label{sec:exp}

\begin{figure}[t]
    \centering
    \includegraphics[trim=0 0 0 17,clip=true,width=0.88\linewidth]{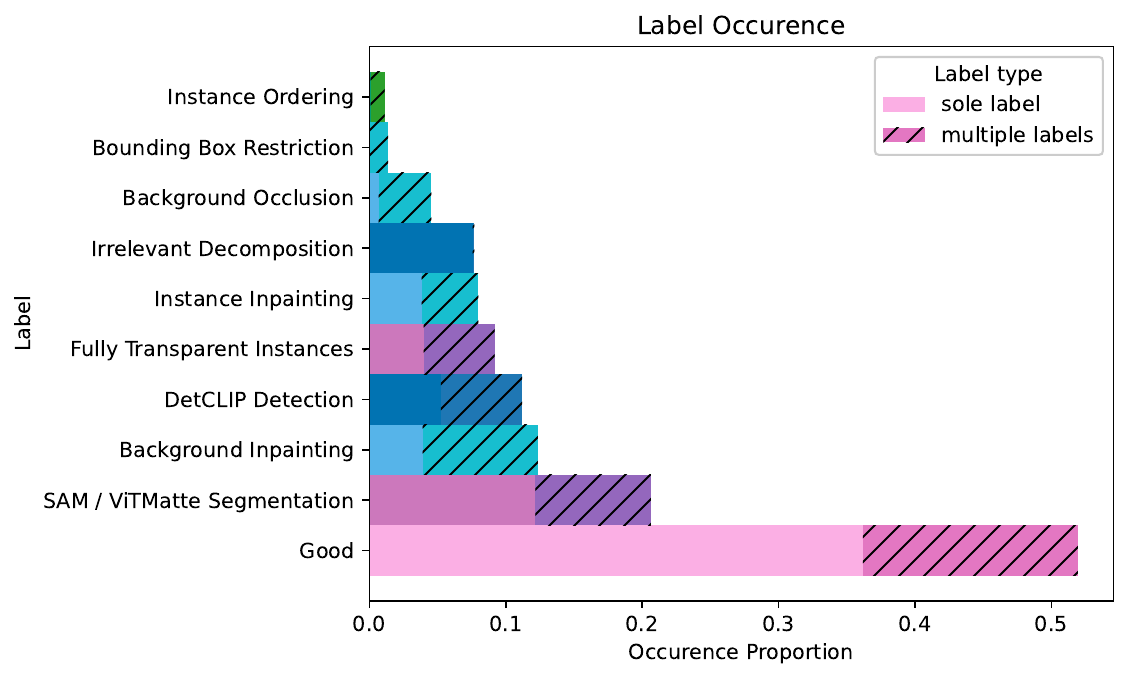}
    \caption{Failure distribution on manually annotated data subset.}
    \label{fig:manual_stats}
\end{figure}

\subsection{Base Datasets}
We run our full method on two datasets that provide sufficient scene compositionality to fully exploit our pipeline: the COCO~\cite{coco} dataset and the Aesthetic V2 6.5 subset of the LAION~\cite{schuhmann2022laion} dataset. The Aesthetic subset filters the complete LAION dataset, selecting only images with an Aesthetic score of at least 6.5 and encompasses 625K images. 
To limit scene complexity and facilitate inspection, we only consider images that comprise one to five instances, which we determine using our object detector's output. We process all COCO images (58K images), and a random subset of 100K LAION images to limit computational cost.

\subsection{Data Curation}
\label{sec:datacur}

We aim to build a dataset comprising high quality decompositions, and exclude potential failure modes. To this end, we manually inspect and label our processed data, identifying six main causes for decomposition failure: 

\begin{figure}[t]
    \centering
    \includegraphics[width=0.88\linewidth]{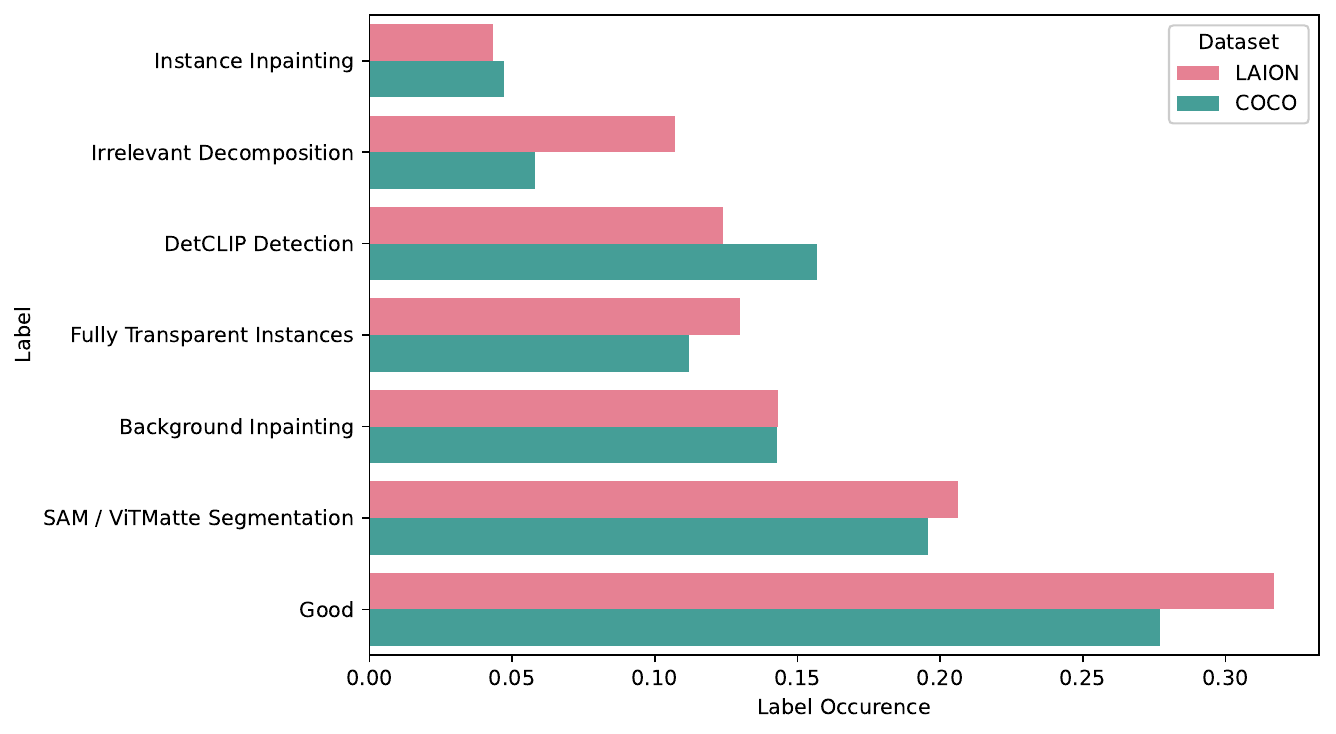}
    \caption{Failure distribution on automatically annotated data.}
    \label{fig:auto_stats}
\end{figure}

\begin{itemize}
    \item \textbf{Object detection}: missing a key instance in an image, or multiple detections of the same object.
    \item \textbf{Segmentation}: incorrect instance segmentation on the original image, or after inpainting.
    \item \textbf{Background inpainting}: erroneous inpainting of the background image. This can be caused by imperfect segmentations, and our pipeline not accounting for causal visual instance effects on the scene (\eg shadows).
    \item \textbf{Instance inpainting}: incorrect or incomplete inpainting of an instance. This often happens due to mask shape or pose biases (\eg person holding a guitar).
    \item \textbf{Truncated instances}: image matting overly eroding the alpha mask of very small instances.
    \item \textbf{Irrelevant decomposition}: scenes that are not suited for instance-wise decompositions (\eg scenes where part of the landscape was incorrectly detected).   
\end{itemize}

Additionally, for analysis purposes, we annotate examples where the instance ordering is incorrect, where background elements occlude instances, and where instance completion is restricted by our bounding box constrained re-segmentation. We provide visual examples of failure modes in Supp.~Materials. Using Voxel FiftyOne \cite{moore2020fiftyone}, we annotate 5000 randomly selected images from our processed pool of LAION Aesthetic $6.5$ images, adding the `good' label for successful decompositions. To mitigate biases, annotations were carried out independently by 3 annotators. We highlight that multiple labels can be assigned to a single image, and notably associate the `good' label with other labels, when defects are minor and do not affect the overall validity of the decomposition.  The distribution of failure modes over this manually annotated set is shown in Fig.~\ref{fig:manual_stats}, highlighting an overall success rate of $36\%$ ($52\%$ with minor defects). We can see that segmentation issues are the biggest failure mode, followed by inpainting and object detection. Failures of our novel ordering, 
together with bounding box restrictions and background occlusion were the rarest issues.

Following~\cite{zhang2023text2layer}, we leverage our manual annotations to train two classifiers to automatically annotate the rest of our processed data: an image level classifier flagging background and irrelevant decomposition issues, and an instance level multi-label classifier identifying remaining failure modes. Details on our classifier architectures and training process are discussed in Supp.~Materials. Fig.~\ref{fig:auto_stats} show the resulting label distribution for both LAION and COCO datasets. We adopt a conservative approach and select images with \emph{only} a confident `good' label as successful decompositions, and report only this portion of `good' labels in Fig.~\ref{fig:auto_stats}. This yields 16K decompositions from COCO, and 28.9k from LAION, for a total of 44.8K annotations in our MuLAn dataset. 
Our automated failure modes distribution for LAION is very similar to our manually annotated portion, with segmentation and inpainting consistently the prominent issues. COCO follows a similar distribution, with larger object detection errors. This is expected as COCO is well known to be a challenging object detection benchmark (with COCO \cite{coco} and LVIS \cite{lvis} annotations), with complex scenes. In contrast, LAION comprises simpler scenes with fewer instances. 

\begin{figure}[t!]
    \centering
    \begin{subfigure}{0.46\linewidth}
    \centering\includegraphics[trim=0 0 0 25,clip=true,width=\linewidth]{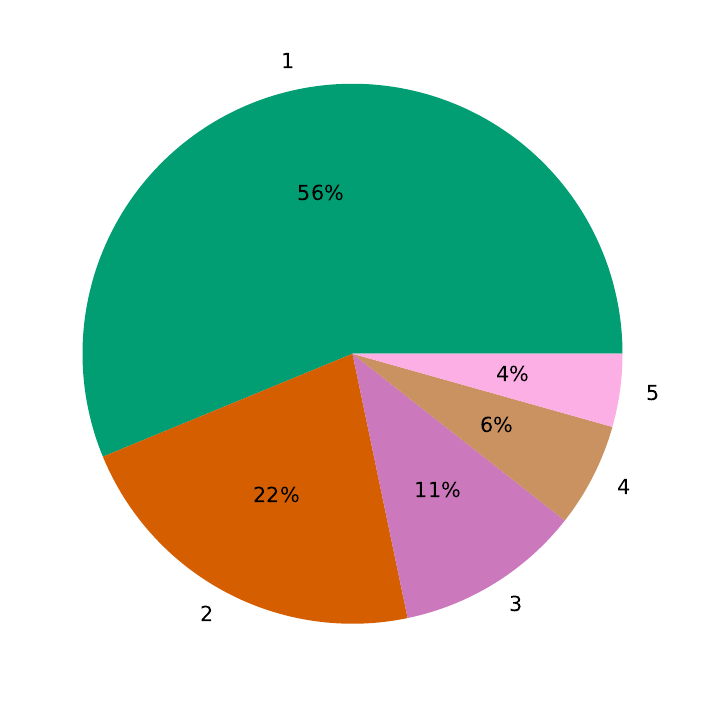}
    \caption{MuLAn-LAION}
    \end{subfigure}%
    \hfill %
    \begin{subfigure}{0.46\linewidth} 
    \centering\includegraphics[trim=0 0 0 25,clip=true,width=\linewidth]{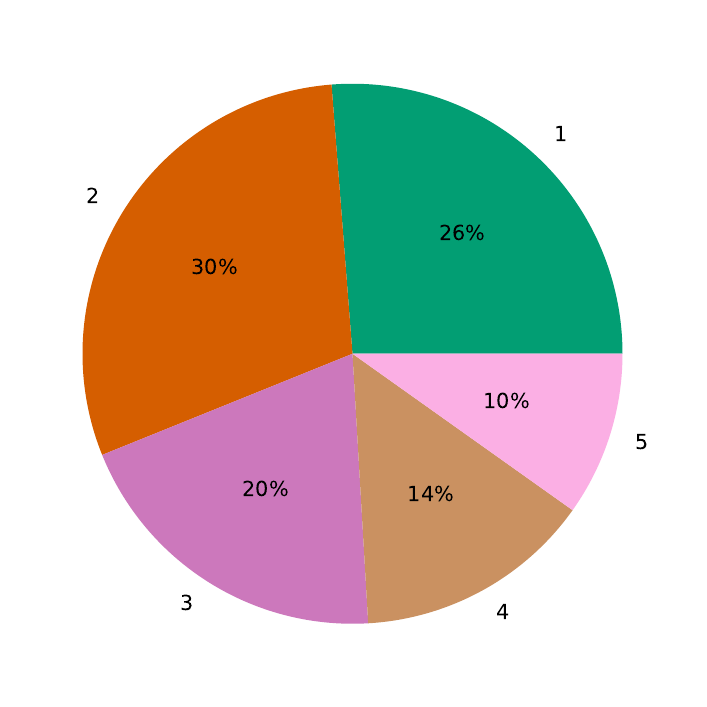}
    \caption{MuLAn-COCO}
    \end{subfigure}%
    \caption{Scene distribution of MuLAn-LAION and MuLAn-COCO datasets. Distribution of number of categories per image.}
    \label{fig:scene_distrib}
\end{figure}

\subsection{Dataset Analysis}
\label{sec:datastats}

With our curated high quality annotations, we further analyse scene distribution and diversity of our 44.8K annotated images.
Fig.~\ref{fig:scene_distrib} shows scene distribution in MuLAn in terms of number of instances per image. We can see that the LAION dataset has a majority of single instance images, which can be linked to the fact that highly aesthetic images tend to be simple scenes (\eg~portraits - this is also highlighted in Supplementary Fig.~\ref{fig:acc_rej_scene_distrib}). Nonetheless, MuLAn-LAION does contain sufficiently complex scenes, with $21\%$ ($\simeq 6$K) of images having more than three instances per images. 
MuLAn-COCO achieves good scene diversity, with $10\%$ of the dataset comprising five instances, almost half of the dataset ($44\% \simeq 7$K) comprising more than three instances and only $28\%$ ($\simeq 4.5$K) of single instance images.

\begin{figure}[t]
    \centering
    \begin{subfigure}{0.47\linewidth}
    \centering\includegraphics[width=\linewidth]{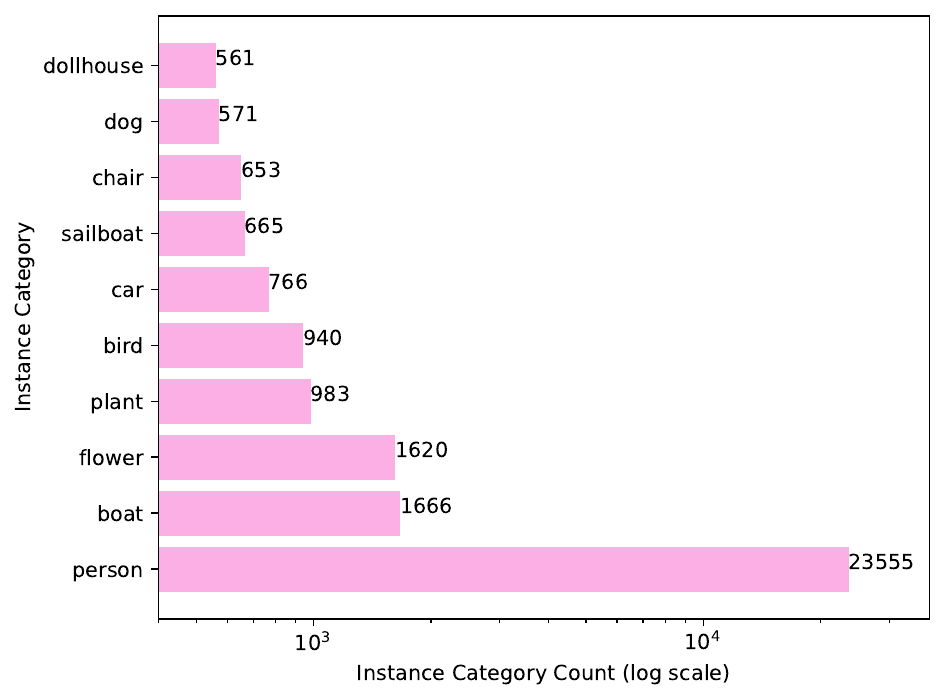}
    \caption{MuLAn-LAION}
    \end{subfigure}%
    \hfill %
    \begin{subfigure}{0.47\linewidth}
    \centering\includegraphics[width=\linewidth]{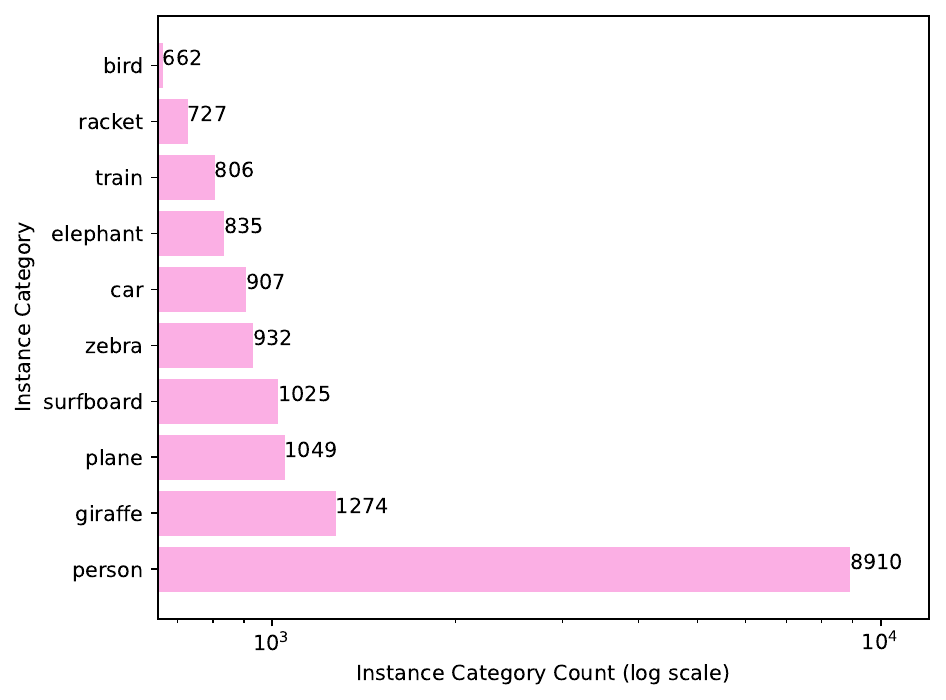}
    \caption{MuLAn-COCO}
    \end{subfigure}
    \caption{Top 10 most common categories in MuLAn subsets.}
    \label{fig:topcat}
\end{figure}

Next, we investigate scene diversity in terms of instance types. 
Out of the $942$ detection categories, we obtain $662$ and $705$ categories in MuLAn-COCO and MuLAn-LAION, respectively, with a total of $759$ categories in MuLAn. Fig.~\ref{fig:topcat} shows the top ten most common categories in each dataset. While the \emph{person} class is the majority class for both, it is overwhelmingly dominant in LAION. Besides persons, MuLAn-LAION mainly comprises inanimate and decor objects, while COCO comprises more active scenes, notably with animals and sports. Of the top ten categories, only three are common to both datasets (person, car and bird). These results highlight the complementarity of both dataset subsets, with MuLAn-LAION focusing on simpler, high quality and visually pleasing scenes, while MuLAn-COCO showcases more diverse scene types. The complete, sorted, list of categories per sub-dataset is available in Supp.~Materials. 

\begin{figure}
    \centering
    \begin{subfigure}[t]{\linewidth}
    \centering\includegraphics[width=0.95\linewidth]{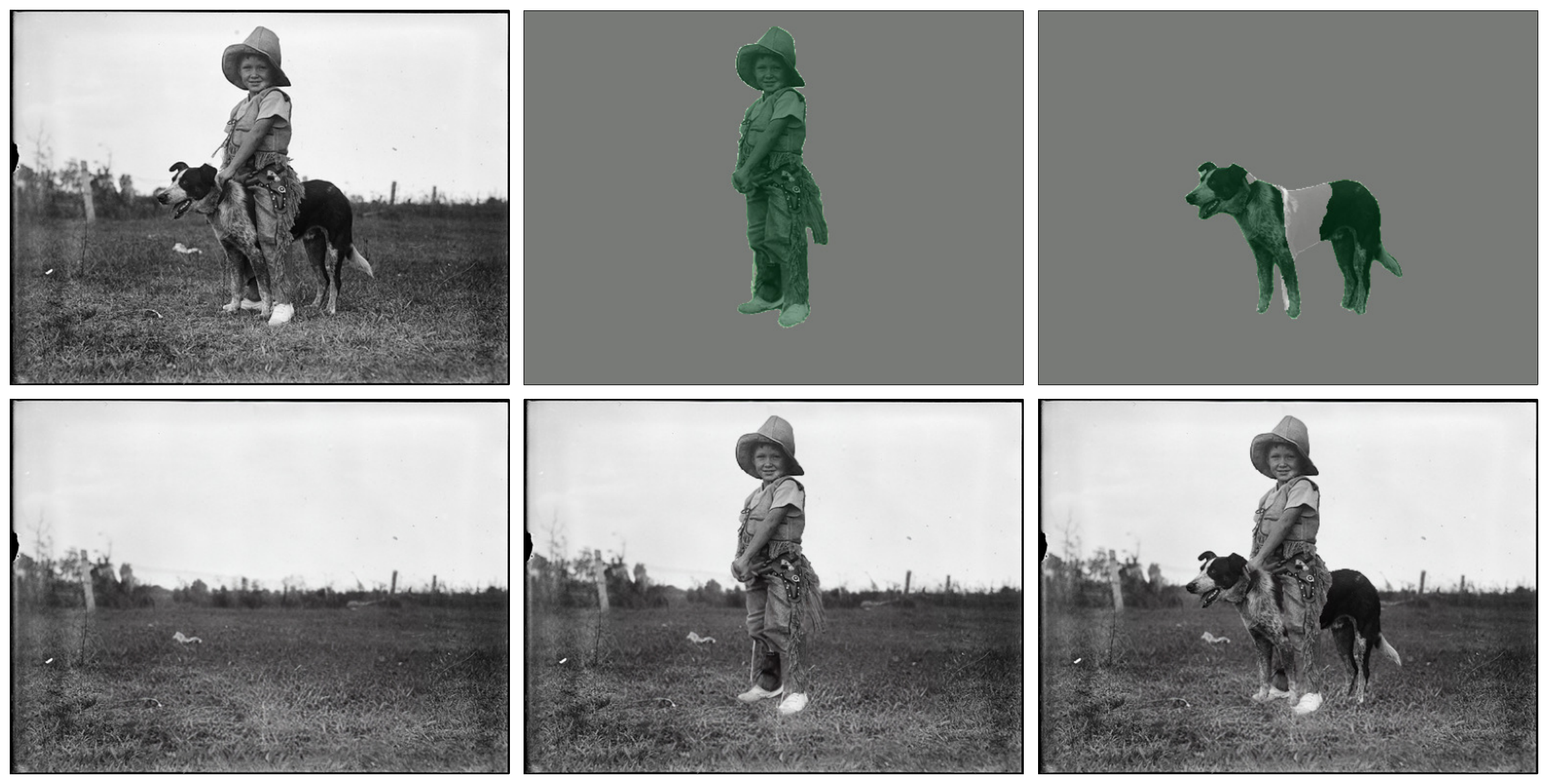}
    \end{subfigure}
    \begin{subfigure}[t]{\linewidth}
    \centering\includegraphics[width=0.95\linewidth]{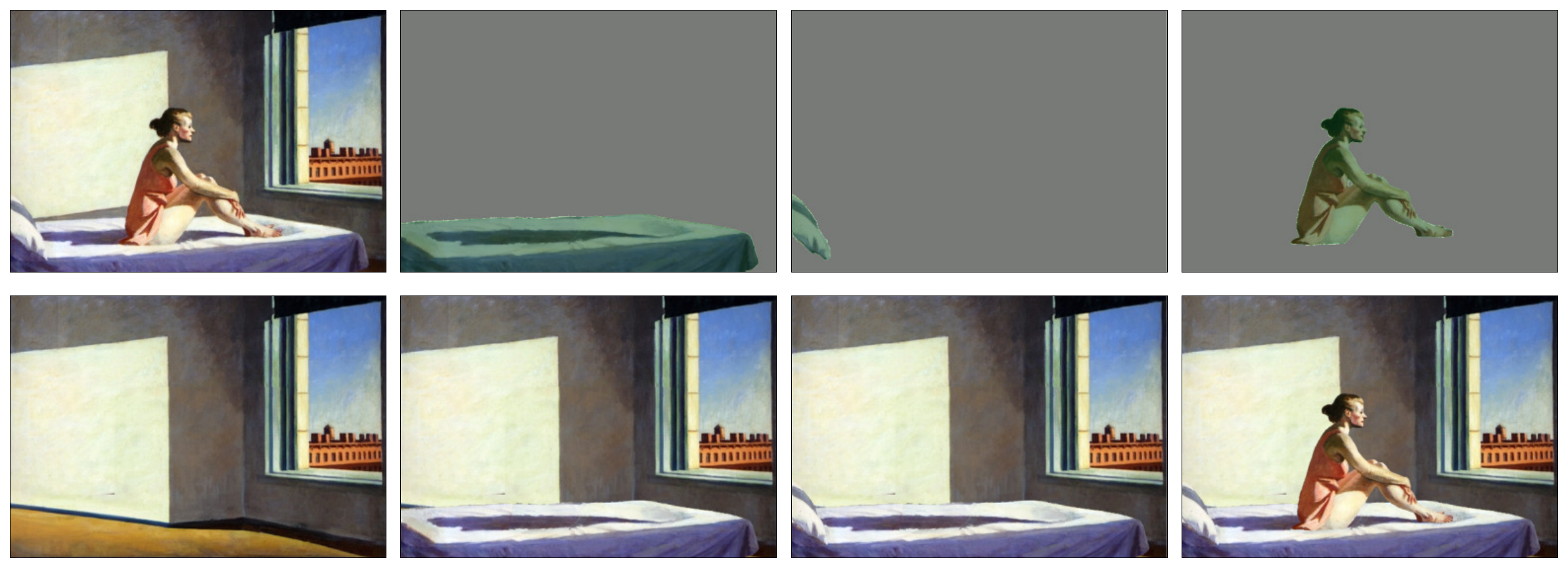}
    \end{subfigure}
    \begin{subfigure}[t]{\linewidth}
    \centering\includegraphics[width=0.95\linewidth]{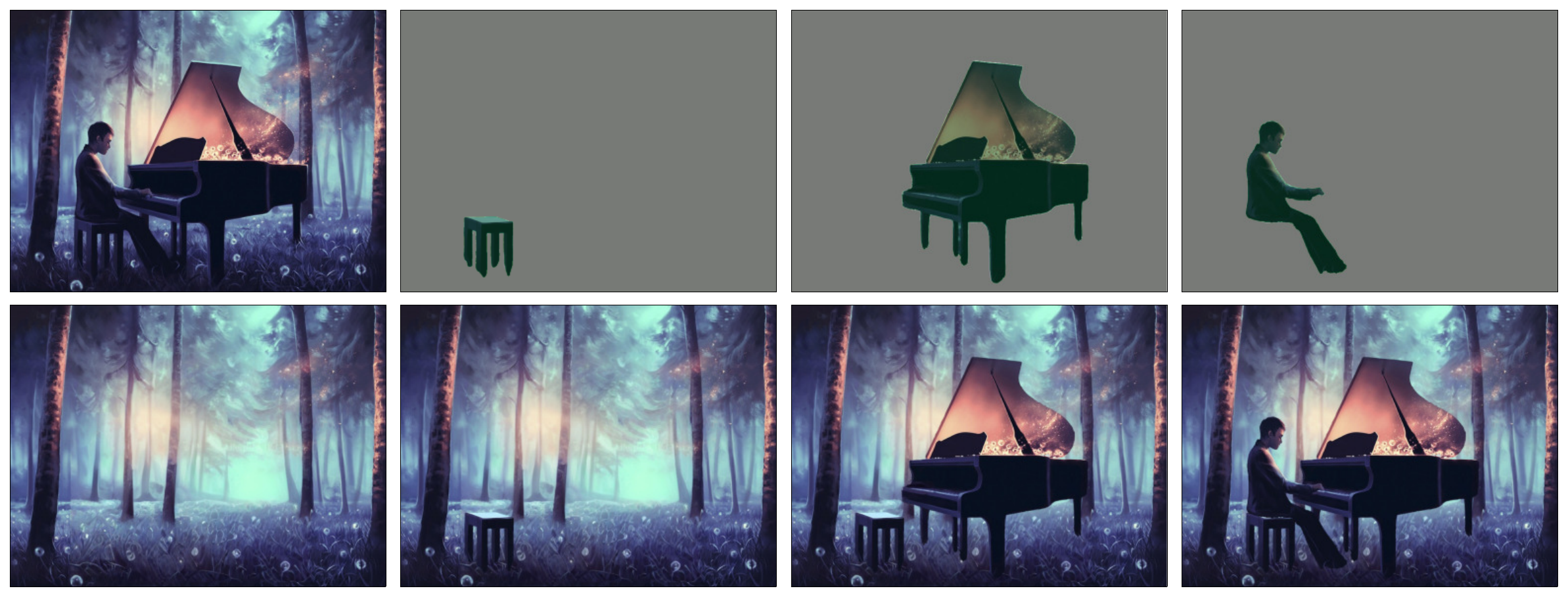}
    \end{subfigure}
    \caption{Visualisation of 3 decompositions from MuLAn-COCO (top) and MuLAn-LAION (bottom 2). From left to right: original image, instance RGBA image with green alpha overlay (top row); reconstructed images by adding layers one by one (bottom row). }
    \label{fig:results}
\end{figure}

Finally, Fig.~\ref{fig:results} presents additional visual examples of RGBA decompositions from MuLAn, showcasing a variety of scene compositions, styles and category types. Additional examples are available in Supp.~Materials.

\subsection{Dataset Applications}
\label{sec:datausages}

To illustrate the potential utility of our MuLAn dataset, we provide two experiments showcasing distinct example scenarios, under which our dataset can be leveraged.

\begin{figure}[t]
    \centering
    \includegraphics[width=0.94\linewidth]{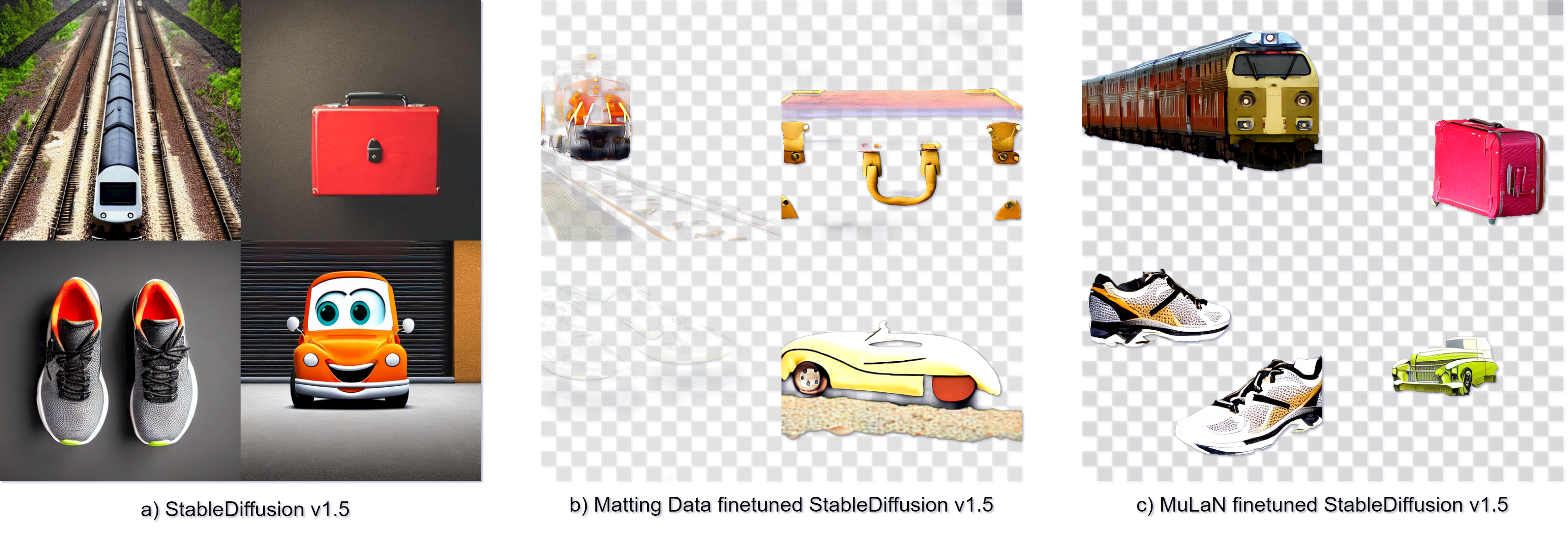}
    \caption{RGBA generation results. Captions: ``a train is approaching``, ``a red suitcase``, ``a pair of running shoes``, ``a cartoon car is parked``. For StableDiffusion, ``on a black background`` was added.}
    \label{fig:rgba_qualitative}
\end{figure}

\noindent \textbf{RGBA Image generation.}
Our first application leverages MuLAn instances to adapt a diffusion model to generation of images with a transparency channel, by finetuning both the VAE and Unet of the Stable Diffusion (SD) v1.5 \cite{rombach2022high} model. 
In Fig~\ref{fig:rgba_qualitative}, we provide visual comparisons of generated images, that are obtained using SD v1.5 with ``on a black background'' appended to the prompt and finetuned on our dataset in comparison with a model finetuned on 15,791 instances from multiple matting datasets. 
We can see that our dataset allows to generate better quality RGBA instances, due to a better understanding of transparency channels.

\begin{figure}[t]
\centering
\begin{subfigure}{0.32\linewidth}
    \centering
    \includegraphics[width=\linewidth]{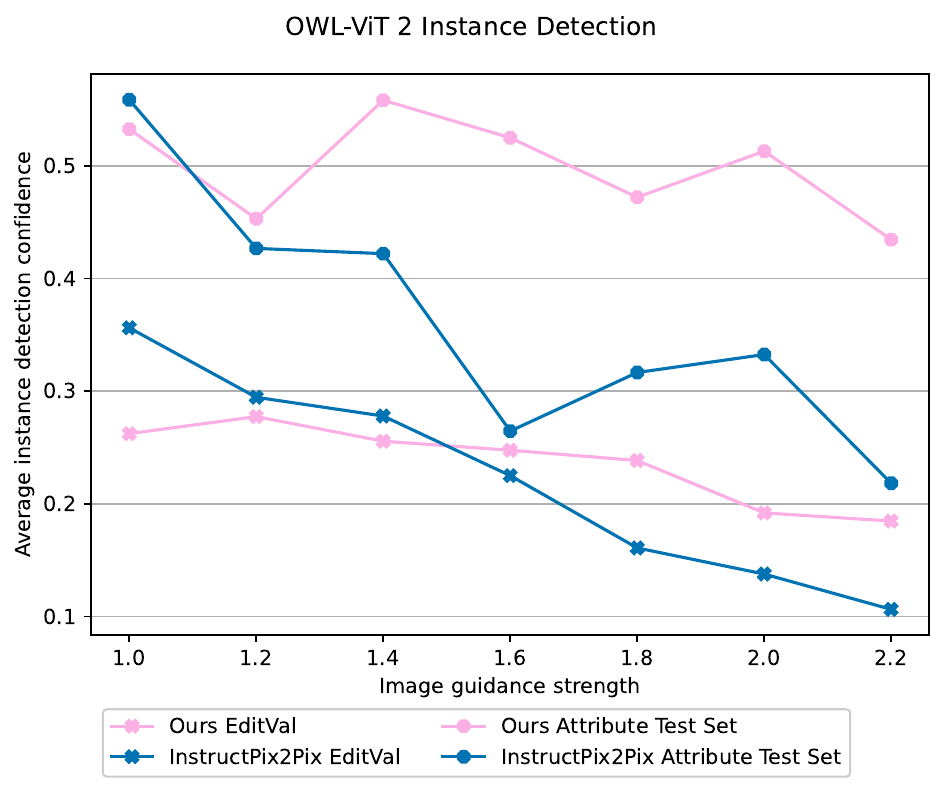}
    \caption{OwlViT 2 edit detection average confidence}
    \label{fig:ip2p_owlvit_detection}
\end{subfigure}%
\hfill%
\begin{subfigure}{0.32\linewidth}
    \centering
    \includegraphics[width=\linewidth]{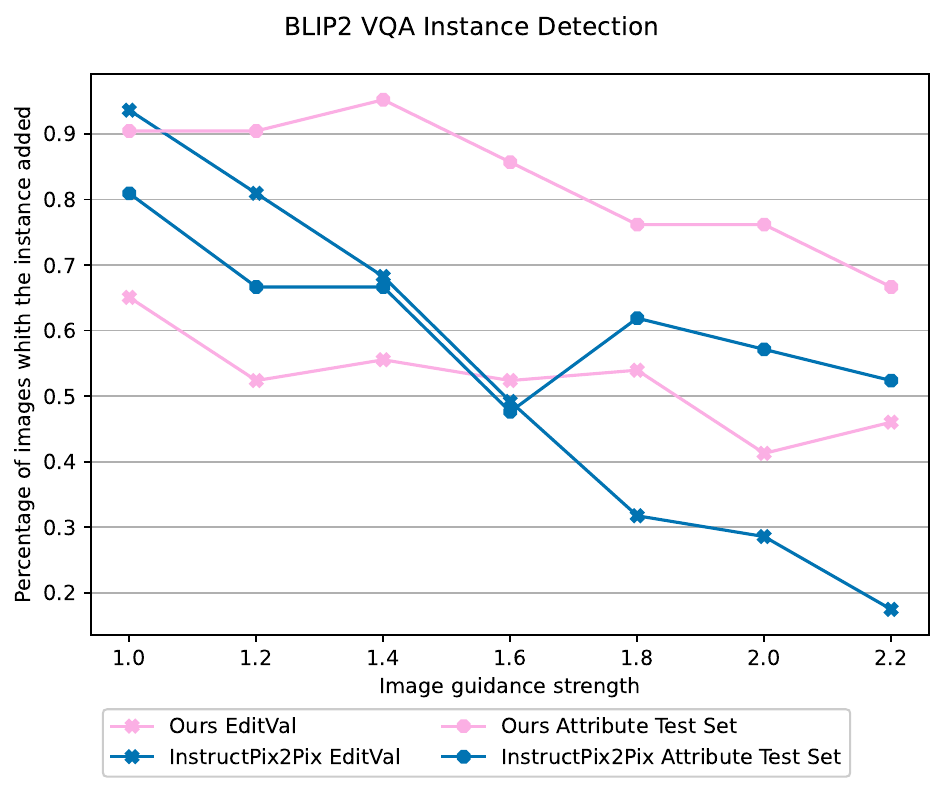}
    \caption{BLIP2 VQA edit detection rate}
    \label{fig:ip2p_blip_detection}
\end{subfigure}%
\hfill%
\begin{subfigure}{0.32\linewidth}
    \centering
    \includegraphics[width=\linewidth]{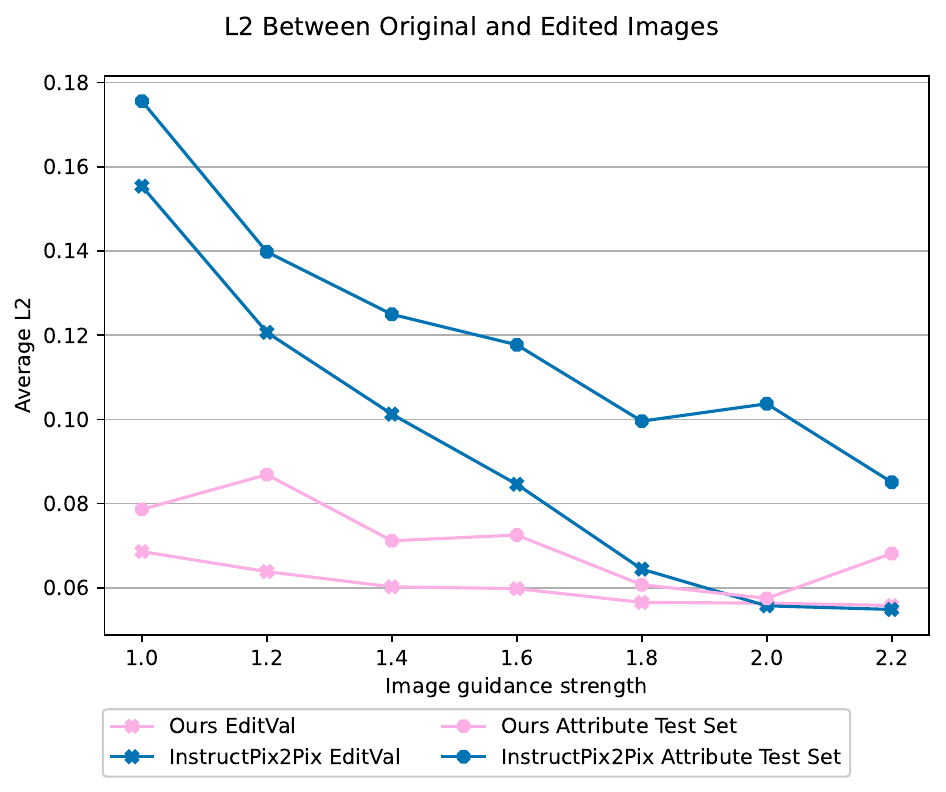}
    \caption{L2 between edited and original images}
    \label{fig:ip2p_l2_edit}
\end{subfigure}
\caption{Instance addition. Quantitative analysis.}
\label{fig:ip2p_quantitative}
\end{figure}

\begin{figure}[t]
\centering
\includegraphics[width=0.94\linewidth]{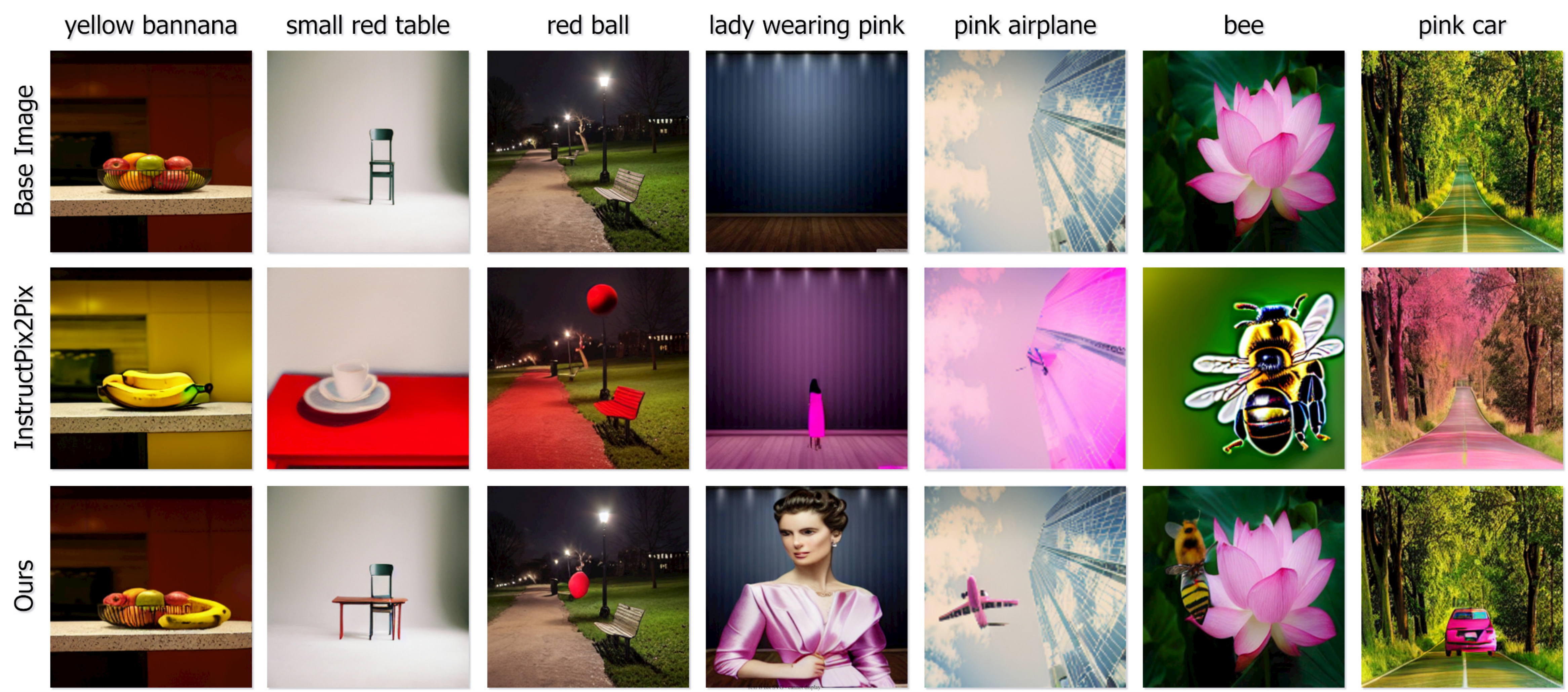}
\caption{Instance addition. Qualitative examples.}
\label{fig:ip2p_qualitative}
\end{figure}

\noindent \textbf{Instance addition.}
Our second application considers an image editing task where the objective is to add instances to an image. We finetune an InstructPix2Pix \cite{brooks2023instructpix2pix} model, taking advantage of our ability to seemlessly add or remove instances in our RGBA stacks. Our training data for InstructPix2Pix comprises
triplets $(I_{S\leq i},I_{S\leq i+1},C_{S_{i+1}})$ 
where $C_{S_{i+1}}$ is the instance caption of layer $i+1$ and $I_{S\leq i}$ is the RGB image obtained by flattening the incomplete RGBA stack 
up to \mbox{layer $i$}. To assess performance, we use EditVal's instance addition evaluation strategy~\cite{basu2023editval}. We report results on the benchmark introduced in~\cite{basu2023editval} (which adds objects without attributes) and build an additional attribute driven evaluation benchmark. Additional details on both evaluation metrics and our benchmark are available in Supp.~Materials. 
Fig.~\ref{fig:ip2p_quantitative} highlights that our model has a better and more consistent performance across the spectrum, in particular with regards to scene preservation. 
This is further evidenced in Fig~\ref{fig:ip2p_qualitative}, where it can clearly be seen that our model has substantially lower attribute bleeding and better background preservation. This can be attributed to our training setup guaranteeing background preservation, in contrast with InstructPix2Pix using Prompt-to-prompt~\cite{hertz2022prompt} editing results.

\section{Conclusions}\label{sec:con}
\looseness=-1
In this work, we introduce MuLAn, a novel dataset for generative AI development that comprises over 44K multi-layer annotation of decomposed RGB images. We built MuLAn by processing images from the LAION Aesthetic 6.5 and COCO datasets using a novel pipeline capable of decomposing RGB images in multi-layer RGBA stacks. MuLAn offers a wide range of scene types, image styles, resolutions and object categories. By releasing MuLAn, we aim to open new possibilities in compositional text-to-image generative research. 
Key to building MuLAn is our image decomposition pipeline. We have provided a detailed analysis of the pipeline's failure modes, notably segmentation, detection and inpainting. Future work will investigate solutions to improve performance and increase MuLAn's size. We can notably leverage the modular nature of the pipeline to introduce better performing models, \eg~segmenters or inpainters. Additionally, the pipeline can be used as a standalone solution to decompose images and facilitate editing with common software. To support this, we additionally investigate human in the loop extensions.


{
\small
\bibliographystyle{ieeenat_fullname}
\bibliography{references}
}

\setcounter{page}{1}
\maketitlesupplementary

\setcounter{equation}{0}
\setcounter{figure}{0}
\setcounter{table}{0}
\setcounter{page}{1}
\renewcommand{\theequation}{S\arabic{equation}}
\renewcommand{\thefigure}{S\arabic{figure}}
\renewcommand{\thetable}{S\arabic{table}}

\section{Data Filtering models}

In order to filter out failed decompositions, we train two classifiers using our 5000 manually annotated decomposition results.
The first classifier takes as input the original image, inpainted background image and background inpainting mask. It is a three-way classifier separating successful decompositions from background inpainting failures and irrelevant decomposition. The second one operates at the instance level, taking as input original image, background inpainting mask, inpainted instance and instance alpha mask; it is designed as a multiclass classifier identifying good decompositions, detection, segmentation and inpainting issues, and truncated instances. Both classifiers are built using a frozen pre-trained EfficientNet B0 backbone \cite{tan2019efficientnet}, with the exception of the first layer which is replaced to handle the different input channel size. The background classifier simply trains a fully connected layer on annotated data using a cross entropy loss. 

For our instance level classifier, we adopt a more complex strategy: our annotations are image level, while issues are often encountered at the level of a single instance. Taking inspiration from Multiple Instance Learning (MIL) approaches for weak supervision \cite{shao2021transmil}, we design a multilabel MIL classification task. Each decomposed image represents a bag of instances, with a set of image level categories (good, segmentation, detection, inpainting, truncated). For a given category, a label of 1 indicates that \emph{at least one} instance in this image has this label, while 0 indicates that no instance has this label. To train this model, we first compute individual instance representations using our EfficientNet backbone, then compute a joint image representation using a self-attention mechanism across all image instances \cite{shao2021transmil}. We then feed this global feature vector to a learnable multi-label classifier and train the model using an image-level cross entropy loss. 

We train both models for 200 epochs with learning rates $2e^{-3}$ (background) and $2e^{-5}$ (instance) for a batch size of 16. Our self attention layer has a single head, and dimension $512$, requiring an additional projection layer at the EfficientNet output. For each classifier, we reserve $20\%$ of annotated data for validation purpose. Due to the class imbalance between successful decomposition and rarer failure modes, we adopt a square root sampling strategy \cite{parisot2022long} to train our background classifier, sampling rare classes more often. Performance of filtering models on the validation set is reported in table \ref{tab:supp:filters}. We report F1 scores, which are more accurate evaluators of imbalanced classification results.

\begin{table*}[ht!]
\centering
\begin{tabular}{c|c|ccccl}
\toprule
Filtering Model  & Success label & \multicolumn{5}{c|}{Reject label}                \\ \midrule
\multirow{2}{*}{Full image classifier} &  & \multicolumn{2}{c|}{Background inpainting}  & \multicolumn{3}{c}{Irrelevant decomposition}                       \\ \cline{3-7} 
& 0.94    & \multicolumn{2}{c|}{0.81}  & \multicolumn{3}{c}{0.81}                  \\ \midrule
\multirow{2}{*}{Instance classifier}   &  & \multicolumn{1}{c}{Segmentation} & \multicolumn{1}{c|}{Inpainting} & \multicolumn{1}{c|}{Truncated instance} & \multicolumn{2}{c}{Detection} 
\\ \cline{3-7} 
 & 0.96  & \multicolumn{1}{c|}{0.88} & \multicolumn{1}{c|}{0.76} & \multicolumn{1}{c|}{}  & \multicolumn{2}{c}{0.90} 
 \\ \bottomrule
\end{tabular}
\caption{F1 scores measuring performance of data filtering models.}
\label{tab:supp:filters}
\end{table*}

\begin{figure}[ht!]
    \centering
    \includegraphics[width=\linewidth]{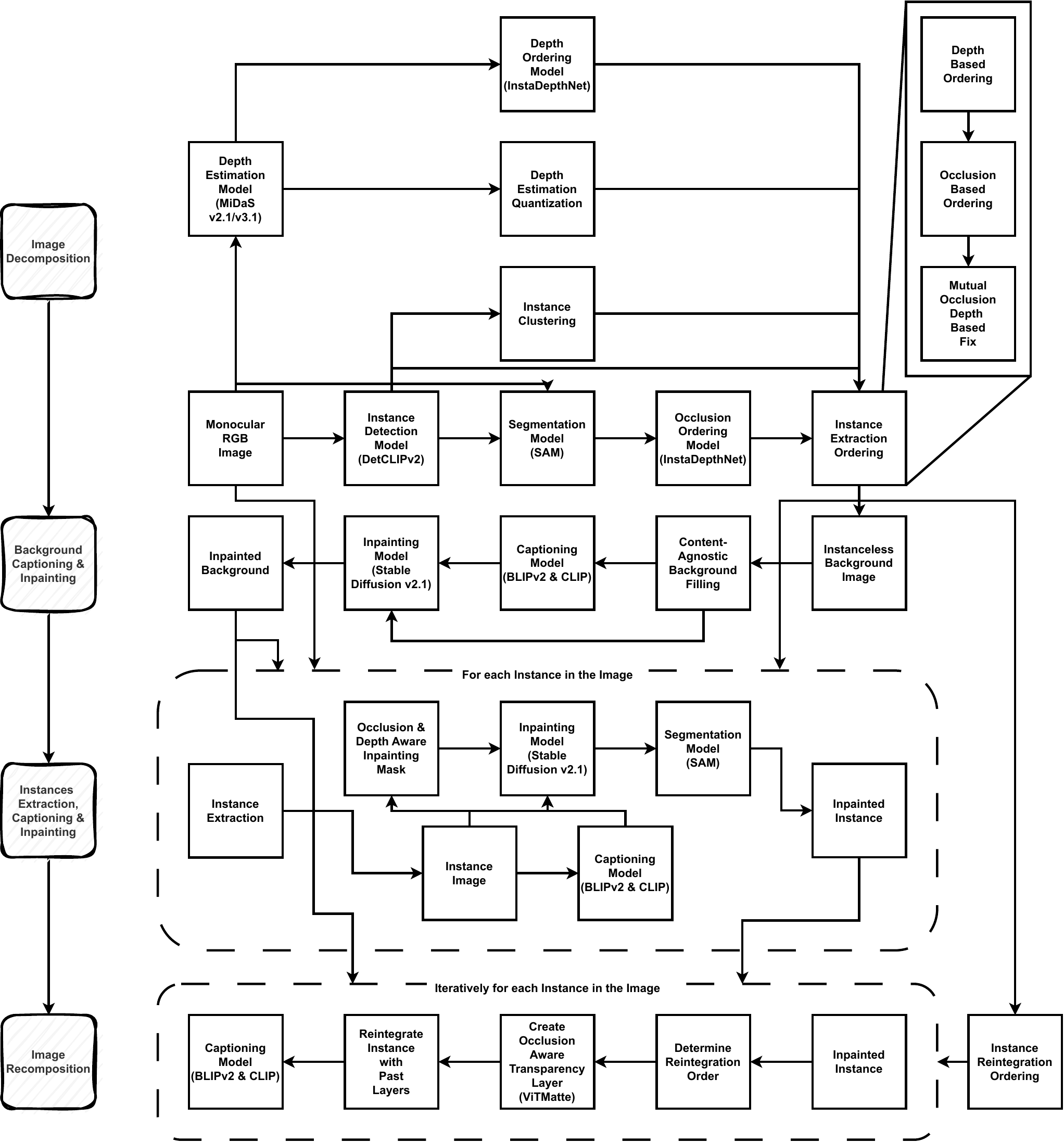}
    \caption{Detailed depiction of the pipeline.}
    \label{fig:detailed_pipeline}
\end{figure}

\section{Pipeline Embodiment}

\subsection{Detailed implementation}

A detailed schematic of our decomposition pipeline is available in Fig. \ref{fig:detailed_pipeline}. We provide additional implementation details below.

\paragraph{Detection.} A reimplementation of DetCLIP v2 \cite{yao2023detclipv2} is used with a SWIN-L backbone. We use an instance score threshold of 0.25, and a Non Maximal Suppression threshold of 0.9. Our class list is attached to this Supplementary Material.

\paragraph{Segmentation.} We use the Segment Anything VIT-h Model \cite{kirillov2023segment} \footnote{Weights found at \href{https://dl.fbaipublicfiles.com/segment\_anything/sam\_vit\_h\_4b8939.pth}{Meta AI Public Files}.} as our segmentation model. We use the DetCLIP v2 bounding box predictions as grounding inputs to the model, and as post-processing, exclude  instances whose largest connected component is smaller than 20 pixels or 0.1 \% of the whole image. This prevents errors in the matting process (TriMap computation) associated with instance pixel counts being too small

\paragraph{Depth Estimation.} MiDaS v3.1 BEiTL-512 \cite{Ranftl2022} \footnote{Weights found on \href{https://github.com/isl-org/MiDaS/releases/download/v3_1/dpt_beit_large_512.pt}{GitHub}.} is used for its robustness and performance. The depth estimation is quantised in bins of 250 relative depth units in order to increase instance separability for the initial instance extraction. 
Images are resized to \mbox{$512{\times}512$} for depth estimation precision.

\paragraph{Instance Occlusion Ordering.} We used the $InstaDepthNet^{o,d}$~\cite{lee2022instance} model to predict both an initial depth ordering as well as an occlusion ordering. The model is build on top of MiDaS v2.0 and takes as input the instance segmentations generated by SAM.

\paragraph{Captioning.} ViT-g FlanT5XL trained via the BLIP-2 paradigm \cite{li2023blip} and further finetuned to produce COCO style captions \cite{coco} \footnote{Weights found on \href{https://huggingface.co/Salesforce/blip2-flan-t5-xl-coco/tree/main}{HuggingFace}.} is used to caption the instances, background and intermediary layers. The captioning prompt that was used is the one used during training \textit{"a photo of"}. In order to increase reproducibility, Beam Search is used with top 32 beams being kept in memory. Besides the 32 predicted captions, the instance's category name and the term ``image'' for the background layer were added as candidate captions. The best caption was then selected using a pretrained CLIP model with VIT-L-patch14 backbone \cite{radford2021learning} \footnote{Weights found on \href{https://huggingface.co/openai/clip-vit-large-patch14}{HuggingFace}.} based on the similarity score between caption and image. Lastly, we used LLaVa  v1.5 7b \cite{liu2023improved} \footnote{Weights found on \href{https://huggingface.co/liuhaotian/llava-v1.5-7b}{HuggingFace}.} to generate detailed captions of the background and the fully recomposed image in order to promote complex captioning based generation \cite{chen2023pixart}. LLaVa was not used on individual instances as it was hallucinating too many details based on the instance's appearance (\eg person's pose). The captioning component takes as input the instance extracted based on the SAM segmentation.

\paragraph{Inpainting.} We use Stable Diffusion v1.5 for our inpainting task. We dilate bounding boxes by a $0.1$ ratio of image size, and crop the input image within the dilated bounding box. This cropped image is used as input to the inpainting model. Inpainting masks are dilated as well using a Guassian blur filter with $\sigma=7$. The difference in sigma is required to guarantee that for the background inpainting instance contents such as hairs are presents. The area within the inpainting mask is filled with a constant value based on the image content \footnote{Based on \href{https://github.com/AUTOMATIC1111/stable-diffusion-webui/blob/4afaaf8a020c1df457bcf7250cb1c7f609699fa7/modules/masking.py}{Stable diffusion webui}.}.

Inpainting is carried out with 50 timesteps. After inpainting, the cropped area is reintroduced and merged with original image pixels according to the inpainting mask. This reduces content degradation, notably from VAE encoding-decoding. To ensure smooth merging, we dilate inpainting masks and soften mask edges using gaussian blur. 

For background inpainting, we use the same prompt across all images: \textit{``an empty scene''} and the following negative prompts: \textit{``complex, text, distortions, poor quality, crowded, non-uniform, item, main subject, large object, foreground object, foreground, heterogeneous, man, woman''}. We additionally append all detected category names in each image to the negative prompts. 

For instance inpainting, we use estimated captions as prompt and the following negative prompts: \textit{``complex, text, poor quality, distortions, crowded, bad anatomy, deformed, missing arms, missing hands, missing legs, extra arms, extra legs, NSFW, nsfw, tiling, bad proportions, cropped, unnatural pose, fused fingers, missing fingers''}. We additionally append all detected category names (that do not pertain to the instance of interest) in each image to the negative prompts. 

\paragraph{Matting.} We use ViT-Matte \cite{yao2023vitmatte} finetuned on Composition 1K \footnote{Weights found on \href{https://drive.google.com/file/d/1mOO5MMU4kwhNX96AlfpwjAoMM4V5w3k-/view?usp=sharing}{Google Drive}.} together with SAM, following the MatteAnything approach~\cite{yao2023matte}. We resize the mask predicted by SAM to \mbox{$256{\times}256$}, and use the dilated bounding box as grounding. Then the mask is both eroded and dilated with a kernel size of 2 for 2 iterations 
in order to automatically generate a TriMap. The dilation and erosion are conservative in order to preserve small instances. We then use ViT-Matte to predict the Alpha channel based on the inpainted instance image and the TriMap. All values below 0.1 are set to 0 to delete the sporadic alpha noise. Then a matting mask is generated from the alpha channel by binarising it. This matting mask is used to extract the matted instance from its inpainted representation. We have mainly done this due to the unreliable nature of inpainting.

\subsection{Instance Ordering}

\paragraph{Algorithm.} We generate our instance ordering in three steps, relying on depth ordering and occlusion information obtained in our decomposition step. First, instances are ordered based on their depth information, from further away to closest (according to instance mean depth value). This can easily be achieved using the instance depth graph, by computing node outdegree: this computes the number of directed edges departing a node, \ie the number of instances that are behind our node. Second, we rely our occlusion graph to refine our ordering: if instance A occludes instance B, instance B will systematically be ordered before instance A. Finally, mutually occluded instances are reordered according to their maximum depth value. In algorithm \ref{alg1}, we provide an algorithmic overview of our instance ordering algorithm, to facilitate reader comprehension. 

\begin{algorithm*}[ht!]
	\caption{Instance ordering procedure}
	\label{alg1}
	\begin{algorithmic}
		\State{{\bf Inputs}: A non sorted list of instances $\mathcal{N}$, \\A list $L^D$ of maximum depth values per instance, \\ A graph $\mathcal{G}^D = (\mathcal{N},\mathcal{E}^D)$ of relative depths, where $e^D_{ij} \in \mathcal{E}^D$ if instance $i$ is in front of instance $j$ (lower depth) \\
        A graph  $\mathcal{G}^O = (\mathcal{N},\mathcal{E}^O)$ of relative occlusions, where 
        $e^O_{ij} \in \mathcal{E}^O$ if $i$ occludes $j$ 
		}
		\State{{\bf Output}: $\mathcal{N}^S$ sorted in inpainting order
		}
        \\
	    \State \textit{Depth based ordering}
        \State \indent Compute node depth outdegree  $\forall n \in \mathcal{N}$:  $outdeg(n) = \sum_i \vert \mathcal{E}^D_{ni} \vert$ \Comment{Number of instances behind $n$} 
        \State \indent $\mathcal{N}^{S} \leftarrow$ Sort $\mathcal{N}$ by ascending outdegree value
        \\
        \State \textit{Occlusion based ordering correction}
        \State \indent \textbf{For} $i=1$ \textbf{to} $\vert \mathcal{N^S}\vert-1$: \Comment{Loop following current order $\mathcal{N}^{S}$}
        \State \indent \indent \textbf{For} $j=i+1$ \textbf{to} $\vert \mathcal{N^S}\vert$: \Comment{Loop through instances inpainted after $i$}
        \State \indent \indent \indent \textbf{if} $e^O_{ij} \in \mathcal{E}^O$ and $e^O_{ji} \notin \mathcal{E}^O$: 
        \State \indent \indent \indent \indent $\mathcal{N}^{S} \leftarrow Swap(i,j)$  \Comment{Swap instances to inpaint occluded instance first}
        \\
        \State \textit{Final adjustment: mutual occlusion}
        \State \indent \textbf{For} $i=1$ \textbf{to} $\vert \mathcal{N^S}\vert-1$: 
        \State \indent \indent \textbf{For} $j=i+1$ \textbf{to} $\vert \mathcal{N^S}\vert$:
        \State \indent \indent \indent \textbf{if} $e^O_{ij} \in \mathcal{E}^O$ and $e^O_{ji} \in \mathcal{E}^O$: \Comment{mutual occlusion}
        \State \indent \indent \indent \indent
        \textbf{if} $L^D_i < L^D_j$: \Comment{instance $j$ is behind $i$ in terms of max depth value}
        \State \indent \indent \indent \indent\indent $\mathcal{N}^{S} \leftarrow Swap(i,j)$  \Comment{Swap instances to inpaint instance that is further away first}
        \end{algorithmic}
\end{algorithm*}

\paragraph{Quantitative analysis.} To assess the performance of our novel ordering strategy, we ablate over its components and use three metrics to evaluate image reconstruction quality: LPIPS \cite{zhang2018unreasonable}, SSIM \cite{wang2004image}  
and MSE.  LPIPS uses deep features across multiple scales from the AlexNet \cite{krizhevsky2012imagenet} to measure the similarity between a perturbed image and the ground truth image, and was shown to align well with human evaluation \cite{zhang2018unreasonable}. 
SSIM is a staple reconstruction metric used across multiple domains. It is the holistic product of three local dissimilarity factors, namely, luminance, variance, and correlation. Finally we used MSE on both the whole image as well the area covered by the background inpainting mask (MSE Masked). All the metrics were calculated only on images with strictly positive bounding box IoUs (\ie~with overlapping instances).

We compare our ordering method to four baselines.  First, \textit{Reverse Decomposition} is the reverse ordering used to extract instances from the image, as detailed in Section 3.1-Instance Extraction. Other baselines are ablations of our ordering strategy and are equivalent to compound effects of the steps from Algorithm \ref{alg1}. \textit{Depth Ordering} sorts instances based on their mean depth value, \textit{+Occlusion Resolution} further adjusts ordering based on occlusion information, \textit{+Mutual-Occlusion Resolution} integrates order correction based on mutual-occlusion information and constitutes the complete ordering process. All reconstructed images have alpha channels that have been predicted by the ViT Matte model. Finally, we further evaluate the impact of our occlusion aware alpha estimation (\textit{+Occlusion Altered Alpha}), where mutually occluded areas are set to be transparent. 

Our depth based ordering achieves the worst reconstruction quality, highlighting the limitations of relying solely on depth information. Second is our reverse decomposition baseline, which relies on clustering and bounding box size heuristics in addition to depth. Replacing these heuristics with occlusion and mutual occlusion based corrections further improve reconstruction quality, with our final ordering approach achieving the top performance. Finally, our occlusion aware alpha estimation further improves image fidelity by a noticeable margin. 
We additionally provide sub-dataset specific results for our complete approach, showing that MuLAn-LAION generally achieves better reconstruction quality than MuLAn-COCO. We attribute this difference to the higher image complexity of COCO dataset when compared with LAION Aesthetic 6.5 (\eg more single instance scenes in MuLAn-LAION).

\begin{table*}[ht!]
\centering
\resizebox{\textwidth}{!}{
\begin{tabular}{l|cccccc}
\toprule
Ordering Logic                       & Masked MAE \textdownarrow      & MAE \textdownarrow             & PSNR \textuparrow               & SSIM \cite{wang2004image} \textuparrow  & LPIPS \cite{zhang2018unreasonable} \textdownarrow  \textdownarrow \\ \midrule 
Reverse Decomposition                & $0.0173_{\pm 0.0244}$          & $0.0157_{\pm 0.0139}$          & $75.9795_{\pm 5.2819}$          & $0.99975_{\pm 0.00058}$                 & $0.0720_{\pm 0.0530}$                             \\ \midrule 
Depth Based                          & $0.0177_{\pm 0.0253}$          & $0.0158_{\pm 0.0142}$          & $75.9391_{\pm 5.3078}$          & $0.99974_{\pm 0.00062}$                 & $0.0723_{\pm 0.0534}$                             \\ 
+ Occlusion Resolution               & $0.0169_{\pm 0.0245}$          & $0.0156_{\pm 0.0139}$          & $76.0262_{\pm 5.2599}$          & $0.99976_{\pm 0.00059}$                 & $0.0713_{\pm 0.0523}$                             \\
+ Mutual-Occlusion Resolution        & $0.0166_{\pm 0.0244}$          & $0.0155_{\pm 0.0138}$          & $76.0552_{\pm 5.2406}$          & $0.99977_{\pm 0.00058}$                 & $0.0711_{\pm 0.0519}$                             \\ 
+ Occlusion Altered Alpha (ours)              & $\mathbf{0.0156_{\pm 0.0229}}$ & $\mathbf{0.0152_{\pm 0.0133}}$ & $\mathbf{76.1859_{\pm 5.1706}}$ & $\mathbf{0.99978_{\pm 0.00054}}$        & $\mathbf{0.0700_{\pm 0.0507}}$           \\ \midrule 
MuLAn - COCO                         & $0.0164_{\pm 0.0175}$          & $0.0221_{\pm 0.0161}$          & $73.9121_{\pm 4.9423}$          & $0.99969_{\pm 0.00068}$                 & $0.0881_{\pm 0.0525}$                             \\ 
MuLAn - LAION                        & $0.0121_{\pm 0.0133}$          & $0.0115_{\pm 0.0079}$          & $77.3296_{\pm 4.4036}$          & $0.99983_{\pm 0.00019}$                 & $0.0665_{\pm 0.0494}$                             \\ 
\bottomrule
\end{tabular}
}
\caption{Evaluation of the instance ordering re-composition on 4400 LAION images. Masked MAE is MAE applied only on the recomposed image region that was inpainted based on the inpainting mask of the background.}
\label{tab:ordering_metrics_laion}
\end{table*}

\section{Dataset details}

\subsection{Usage}
MuLAn is split into two subsets based on the original datasets that are annotated. MuLAn-COCO consists of 16,034 images with 40,335 instances while MuLAn-LAION consists of 28,826 images with 60,934 instances, for a sum total of 44,860 images with 101,269 instances.

\paragraph{Additional dataset statistics}

\begin{figure}[t]
    \centering
    \begin{subfigure}[t]{0.48\linewidth}
    \centering\includegraphics[width=\linewidth]{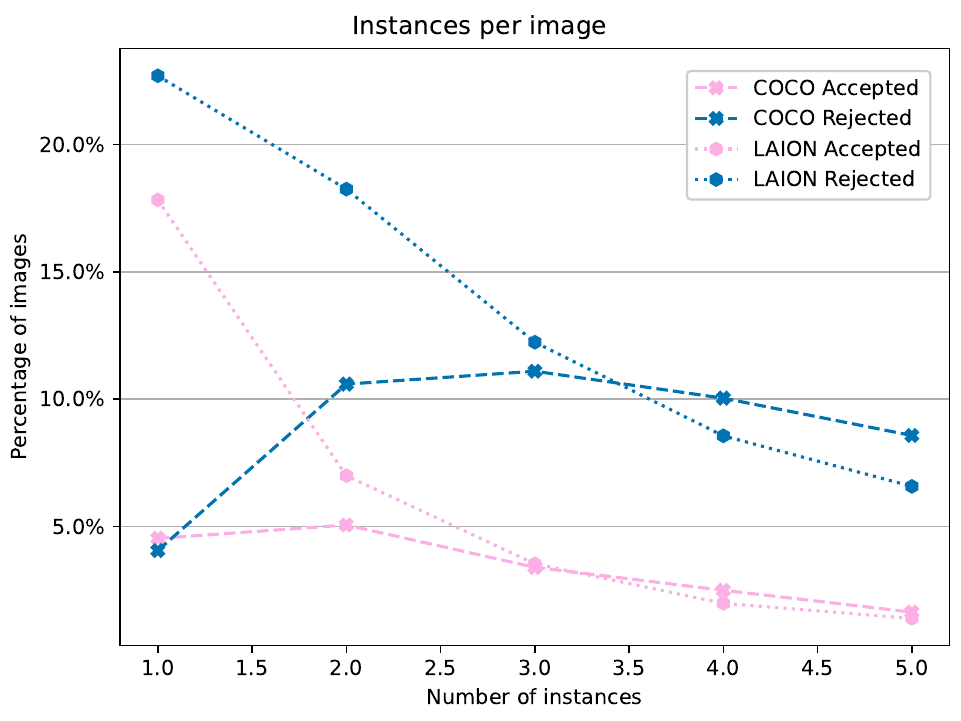}
    \end{subfigure}%
    \hfill %
    \begin{subfigure}[t]{0.48\linewidth}
    \centering\includegraphics[width=\linewidth]{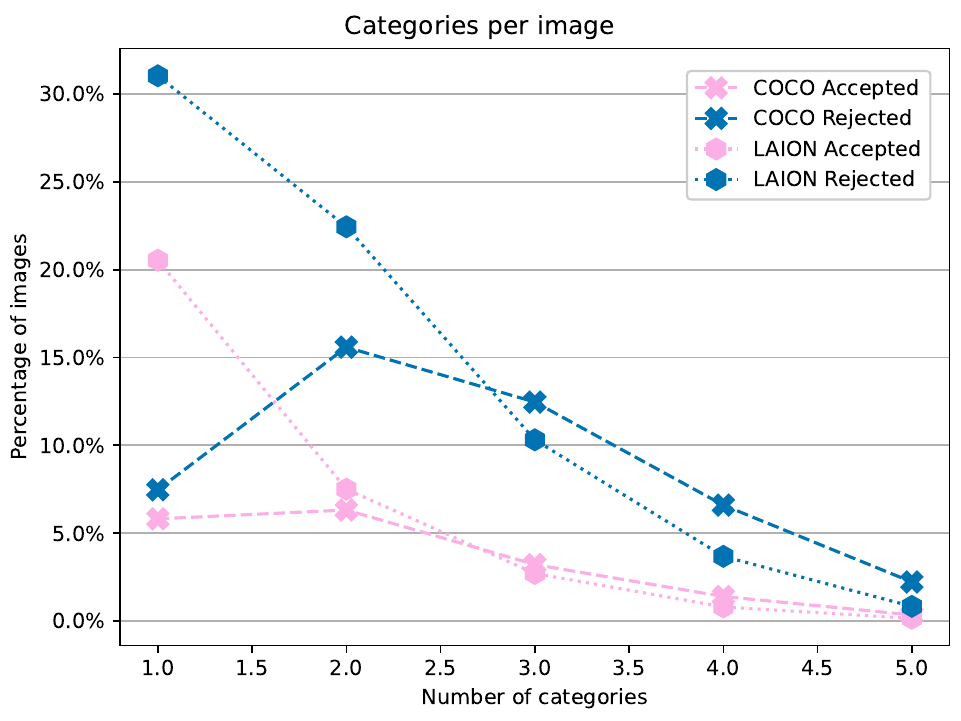}
    \end{subfigure}
    \caption{Scene distribution for successful and rejected decompositions, for both COCO and LAION datasets. Percentage of accepted/rejected decompositions with respect to number of instances (left), and number of categories (right) per image.}
    \label{fig:acc_rej_scene_distrib}
\end{figure}

 Fig. \ref{fig:acc_rej_scene_distrib} shows the percentage of successful and failed decompositions with respect to the number of categories and instances in an image. Differences in scene composition between both datasets are highlighted in this figure: we can see that LAION has a much larger distribution of simpler scenes (1-2 instances and categories), while COCO has a more balanced distribution. In both cases, percentages of successful decompositions decrease as the number of instances increase, highlighting the challenge of handling the intricacies of complex scenes. 

\begin{figure*}
    \centering
    \includegraphics[width=\linewidth]{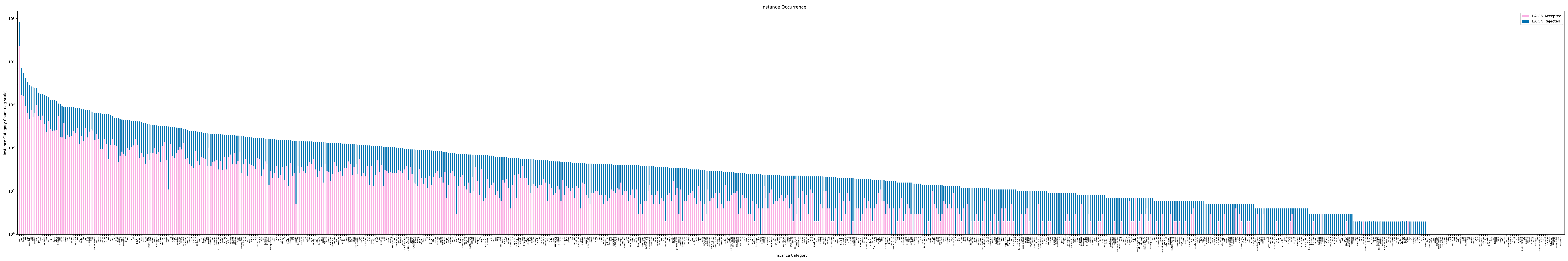}
    \caption{Distribution of all instance categories in the MuLAn-COCO dataset.}
    \label{fig:allcat_COCO}
\end{figure*}

\begin{figure*}
    \centering
    \includegraphics[width=\linewidth]{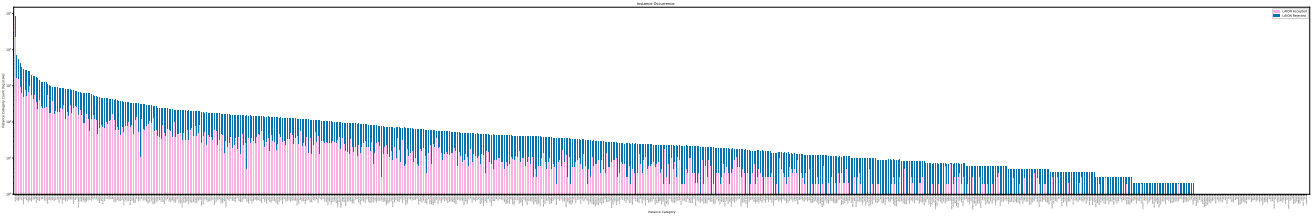}
    \caption{Distribution of all instance categories in the MuLAn-LAION dataset.}
    \label{fig:allcat_LAION}
\end{figure*}

The complete distribution of categories in our dataset is available in Fig.~\ref{fig:allcat_COCO} (MuLAn-COCO) and Fig.~\ref{fig:allcat_LAION} (MuLAn-LAION). These figures additionally highlight which categories have the highest success and failure rates, showing the proportion of accepted and rejected examples.
 
\begin{figure}
    \centering
    \includegraphics[width=0.55\linewidth]{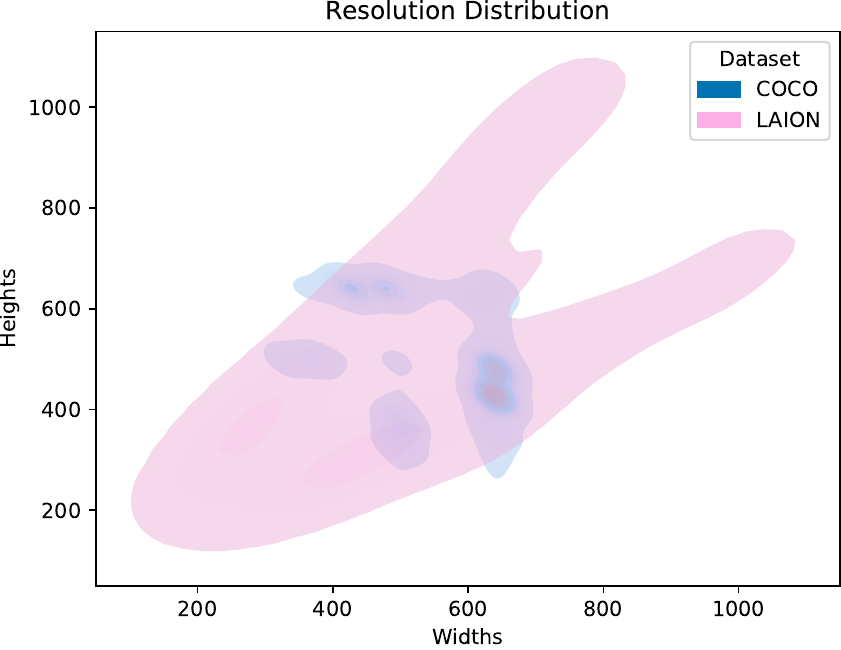}
    \caption{Distribution of image resolutions in MuLAn-COCO and MuLAn-LAION.}
    \label{fig:resolutions}
\end{figure}

In addition, we evidence robust performance and an ability to generalise across a wide-range of image resolutions and qualities as shown in Fig.~\ref{fig:resolutions}, where we report the distribution of image resolutions in our dataset. 

\subsection{Format}
The annotation files are inspired 
by the original COCO dataset annotations, developed by the research community. 
The annotation files contain a dictionary with metadata and required elements to generate the MuLAn dataset given the original images and the annotation file. We highlight that we do not release original image content, and that the decomposed images cannot be reconstructed without access to the original data. The dictionary's contents are outlined in \mbox{Listing~\ref{lst:annot}.}

\paragraph{Layers}
Layers are indexed from 0 to $N$ with Layer 0 being the background. For all layers we have released the masks required to extract the original content from the original image and the inpainted content that needs to be added to the extracted one in order to obtain the layer.

\paragraph{Captions}
Each individual layer comes with a COCO style caption generated by the FLAN T5-XL BLIP 2 style trained model that was further finetuned on COCO styles \cite{li2023blip}. We note that those captions were selected by the CLIP model \cite{chung2022scaling} from 32 candidate captions together with ``image'' or instance tag as default candidates.

For the background (Layer 0), the original image and the recomposed image we have also released captions generated by the LLaVa model in order to encourage the development of generative models based on detailed natural language captions \cite{chen2023pixart}. The instances were not captioned with LLaVa due to its ability to infer details that were not visible in the instance itself but could be found in the original image. We attribute this to the bias resulting from the position of the instance and the gaps of cotent resulted from isolating the instane to be captioned.

\begin{lstlisting}[float=*,caption={Description of our released annotations for a given decomposed image.},label={lst:annot}]
"annotation" : {
    "captioning": {
        "llava": LLaVa model details
        "blip2": BLIP 2 model details 
        "clip": CLIP model details 
    }
    "background": {
        "llava": Detailed background LLaVa caption
        "blip2": COCO style BLIP 2 background caption chosen by CLIP
        "original_image_mask": Original image background content mask
        "inpainted_delta": Additive inpainted background content 
    }
    "image": {
        "llava": Detailed original image LLaVa caption
        "blip2": COCO style BLIP 2 original image  caption chosen by CLIP.
    }
    "instances": {
        "blip2": COCO style BLIP 2 instance caption chosen by CLIP.
        "original_image_mask": Original image instance content mask
        "inpainted_delta": Additive inpainted instance content
        "instance_alpha": Alpha layer of the inpainted instance
    }
}
\end{lstlisting}

\section{Dataset applications details}

\paragraph{RGBA Image generation.} The matting datasets that have been used in finetuning the baseline SD v1.5 are outlined in Table \ref{tab:matting}. We make use of $7$ publicly available matting datasets for a total 15,791 images (\vs 101,269 in MuLAn).
\begin{table}[]
\centering
\resizebox{\linewidth}{!}{%
\begin{tabular}{c|ccc}
\hline
Dataset        & Type   & Resolution       & No. Instances \\ \hline
AIM-500 \cite{AIM}       & Object & $1397\times1260$ & 500           \\
AM-2K  \cite{AIM}        & Animal & $1471\times1195$ & 484           \\
HIM-2K \cite{sun2022instmatt}         & Human  & $1823\times1424$ & 830           \\
RWP-636 \cite{yu2020mask}       & Human  & $1038\times1327$ & 636           \\
PPM-100  \cite{MODNet}      & Human  & $2997\times2875$ & 100           \\
Composition 1K \cite{xu2017deep} & Varied & Varied           & 481           \\
UGD-12K \cite{fang2022user} & Human  & $357\times317$   & 12760        
\end{tabular}%
}
\caption{Matting datasets used to train our RGBA generation baseline.}
\label{tab:matting}
\end{table}

\paragraph{Instance addition.} In order to assess the presence of the new instance we used OWL-ViT 2 \cite{minderer2022simple} similarly to the strategy proposed by EditVal \cite{basu2023editval} for open-set instance detection, and BLIP-2 \cite{li2023blip} for visual question answering. For OWL-ViT 2 we report average detection confidence of the detection (in contrast with EditVal's binary scores). For BLIP-2, we report the percentage of images where the model's answer starts with ``yes" to the following prompt: \textit{``Question: Answer with yes or no, is there a [\emph{instance description}] in the image? Answer:"}. Following the InstructPix2Pix \cite{brooks2023instructpix2pix} evaluation, we keep the text guidance scale constant at $7.5$ and vary the image guidance scale between $1.0$ and $2.2$. Since the EditVal instance addition edits do not include attributes, we created, following the EditVal format, an additional evaluation set of 21 edits where instances to add have attributes (e.g. instance colour). This Attribute Test Set can be found on the project website \footnote{\url{https://MuLAn-dataset.github.io/}}. 

\section{Choice of dataset}

We chose to develop MuLAn based on LAION and COCO datasets due to their pervasiveness within both the generative modelling 
and computer vision communities. The LAION Aesthetic 6.5 subset was specifically chosen due to an appealing 
compromise between cardinality, instance density, scene style \& content, and image quality. Due to 
ethical concerns around fair-use of copyrighted content, we do not include content from original images, and release only our results in annotation format, similar to COCO-based datasets. As such, our data cannot be reconstructed without rightful access to the original image content. 

\section{What did not work and \emph{why}}

In this section we 
exhaustively enumerate 
alternative approaches that were 
explored during our development and the reasons we have chosen the current implementation over the discussed alternatives. 

\paragraph{Improvements to the SAM model: SAM-HQ and SEEM}
We investigated alternatives to the SAM model, which is our main cause of decomposition failures. We first considered SAM-HQ~\cite{samhq}, a model advertised as a global improvement to SAM, notably capable of segmenting details (\eg elongated object, wires) more accurately. While we did observed improvements and more precise segmentations for this type of objects, we observed that SAM-HQ also had a higher tendency to oversegment, leading to potentially reduced instance extraction accuracy. 
We additionally considered the SEEM model~ \cite{SEEM}, which was particularly attractive due to its ability to ground the segmentation using category names. The model was however trained on COCO classes, and we observed reduced performance, compared to SAM, for other categories. 

\paragraph{Grounded-SAM.}
We initially considered Grounded-SAM to extract and segment instances. Grounded-SAM combines Grounding-Dino \cite{liu2023grounding}, SAM \cite{kirillov2023segment} and Tag2Text \cite{huang2023tag2text} in order to predict the instance tags, their bounding box and segmentations. The sequence of three models (vs.~two in our final pipeline version) introduced additional unstability. 
Notably, we used Tag2Text, an image tagger and caption predictor, to identify instances in the images, and provided this list of tags to the Grounding DINO detection model. We observed however that Tag2Text's performance, while very high, was not accurate enough to optimise Grounding DINO's detection performance. In contrast to DETCLIP, Grounding DINO requires the exact list of instances present in the image, and wrong tags can lead to detection of non-existent instances. Ultimately, relying on a single model to detect and identify instances in an image yielded a more robust and consistent performance.

\section{Failure modes: visual examples}

\begin{figure}[t]
    \centering
    \begin{subfigure}[t]{\linewidth}
    \centering\includegraphics[width=\linewidth]{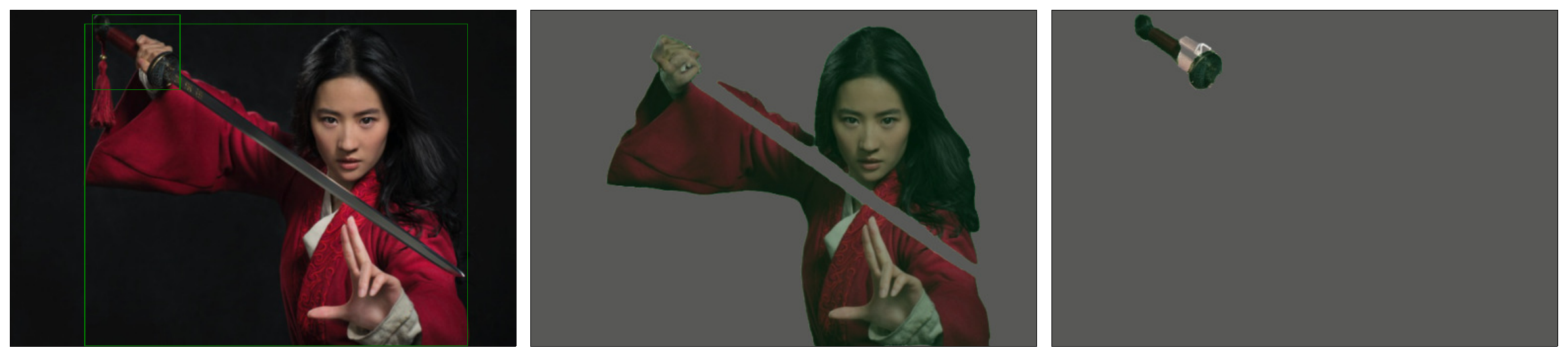}
    \end{subfigure}
    \begin{subfigure}[t]{\linewidth}
    \centering\includegraphics[width=\linewidth]{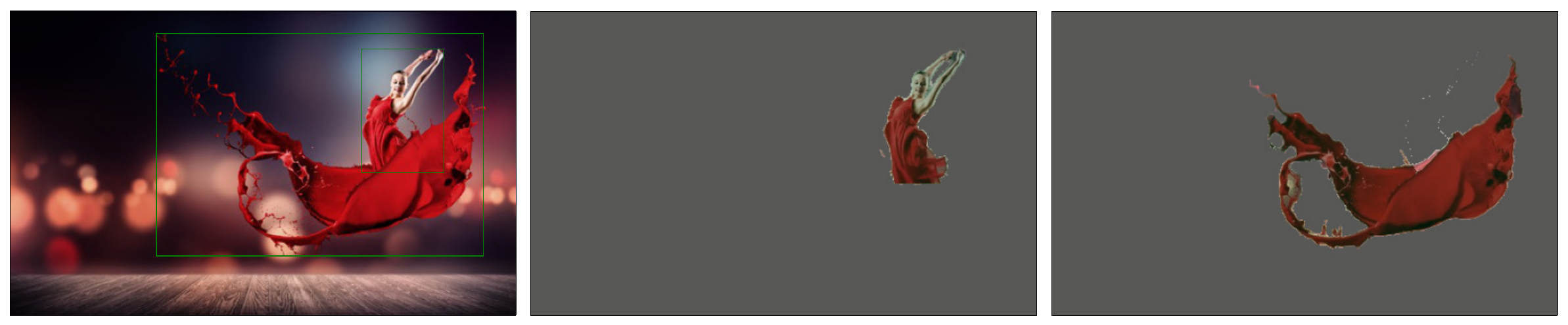}
    \end{subfigure}
    \begin{subfigure}[t]{\linewidth}
    \centering\includegraphics[width=\linewidth]{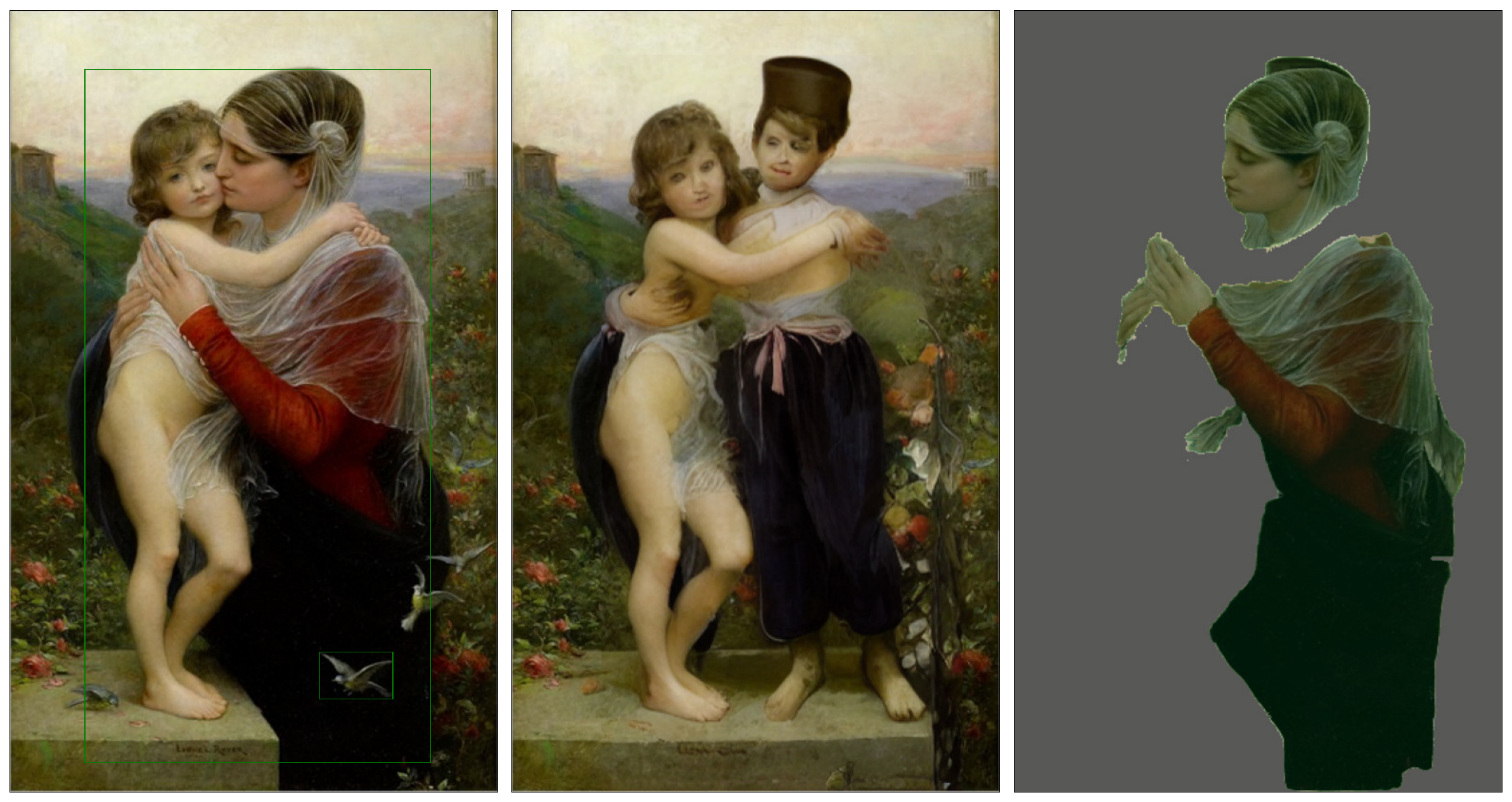}
    \end{subfigure}
    \caption{Visualisation of failure modes: object detection. We show detected instance bounding boxes and affected instances/background image.}
    \label{fig:detclipfail}
\end{figure}

\begin{figure}[t]
    \centering
    \includegraphics[width=\linewidth]{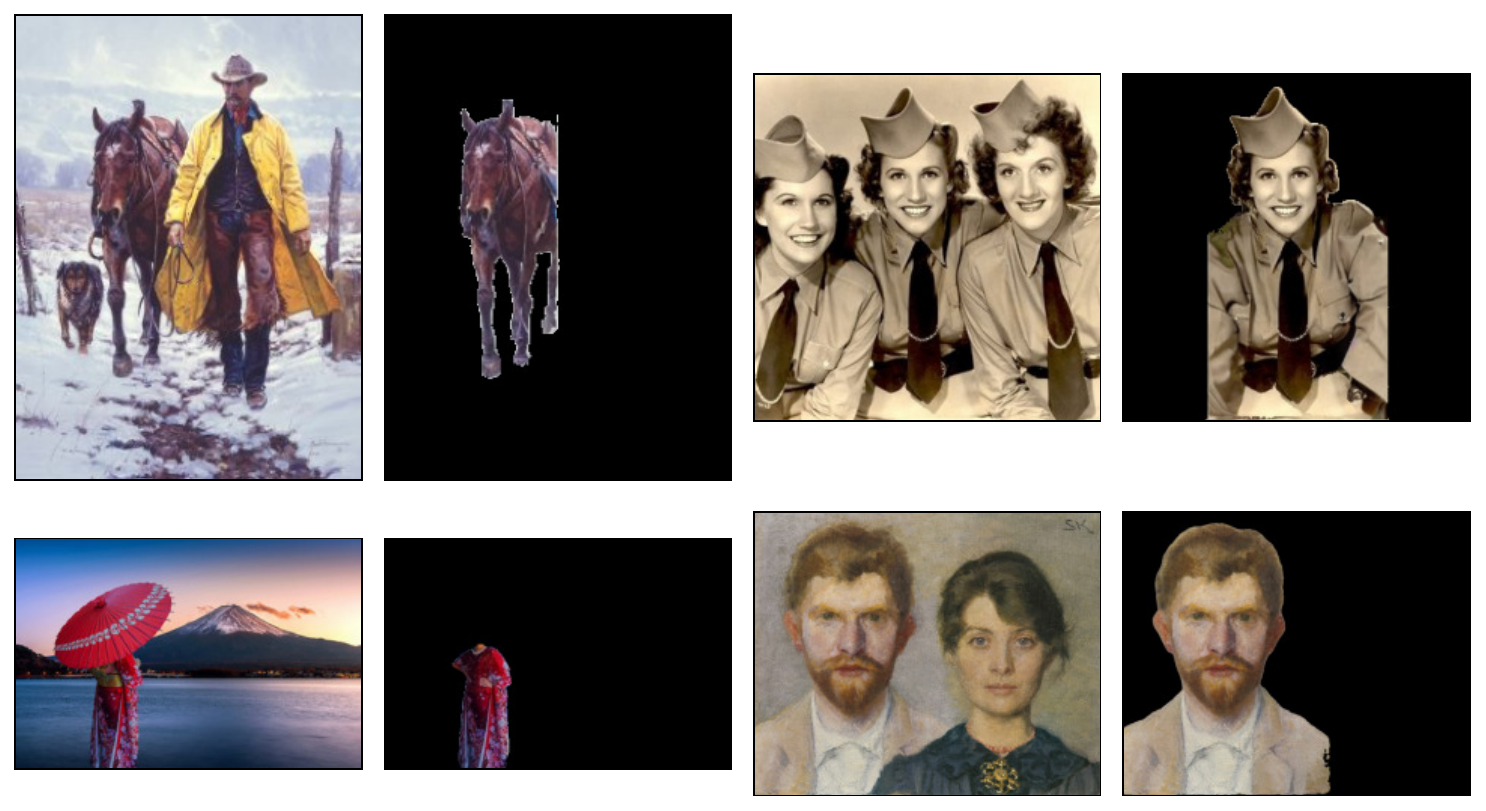}
    \caption{Visualisation of failure modes: bounding box restricted instance completion. Left-right image pairs: original image-instance with failure.}
    \label{fig:bbfail}
\end{figure}

\begin{figure*}[t]
    \centering
    \includegraphics[width=\linewidth]{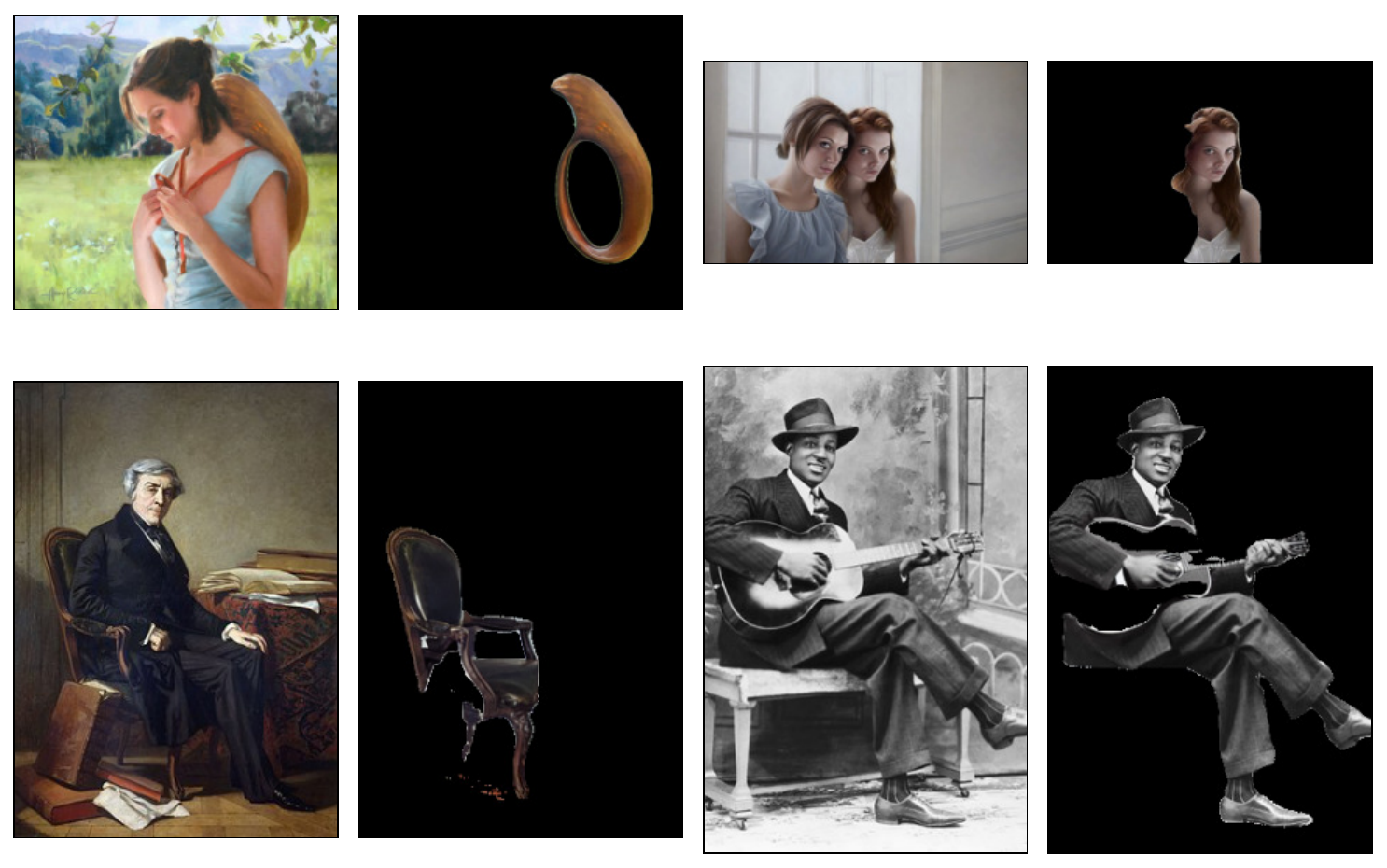}
    \caption{Visualisation of failure modes: instance inpainting. Left-right image pairs: original image-instance with failed inpainting.}
    \label{fig:inpaintingfail}
\end{figure*}

\begin{figure*}[t]
    \centering
    \includegraphics[width=\linewidth]{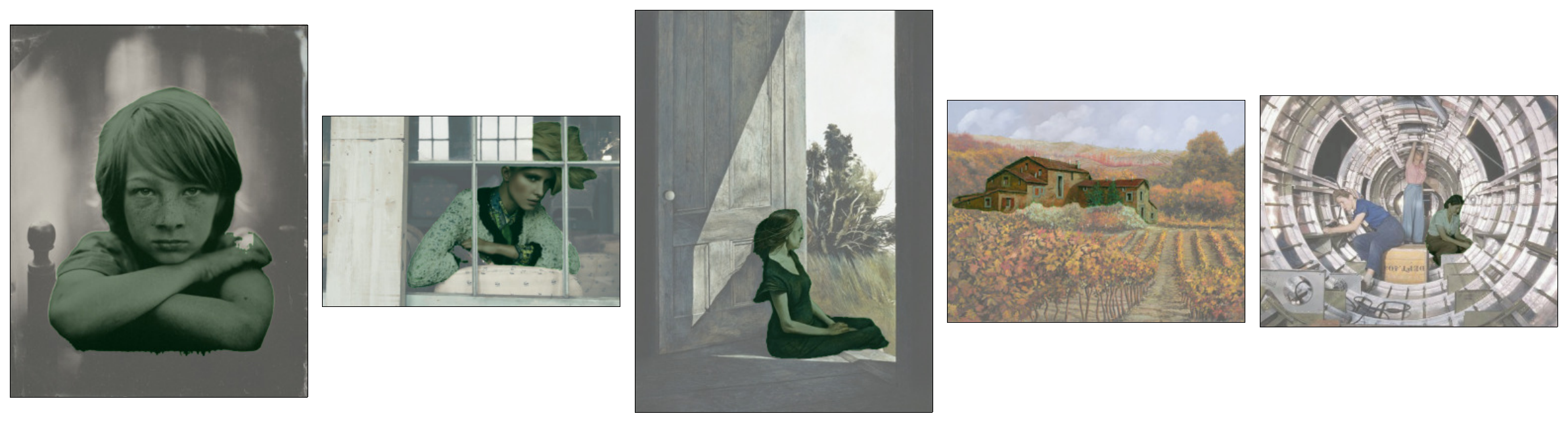}
    \caption{Visualisation of failure modes: background occlusions. Green overlay is the estimated instance segmentation.}
    \label{fig:boccfail}
\end{figure*}

\begin{figure*}[t]
    \centering
    \includegraphics[width=\linewidth]{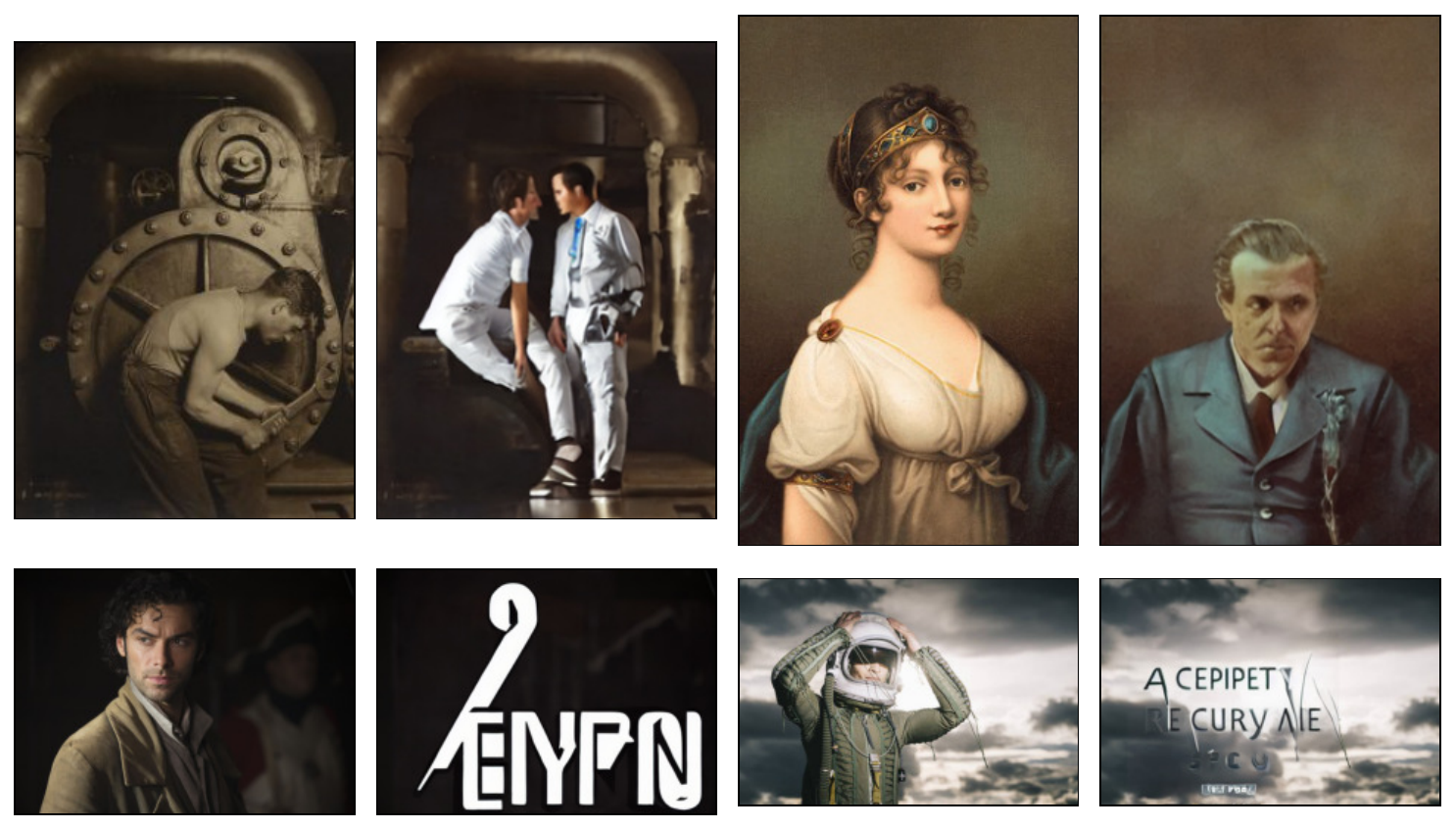}
    \caption{Visualisation of failure modes: background inpainting. Left-right image pairs: original image-inpainted background}
    \label{fig:bgfail}
\end{figure*}

\begin{figure*}[t]
    \centering
    \includegraphics[width=\linewidth]{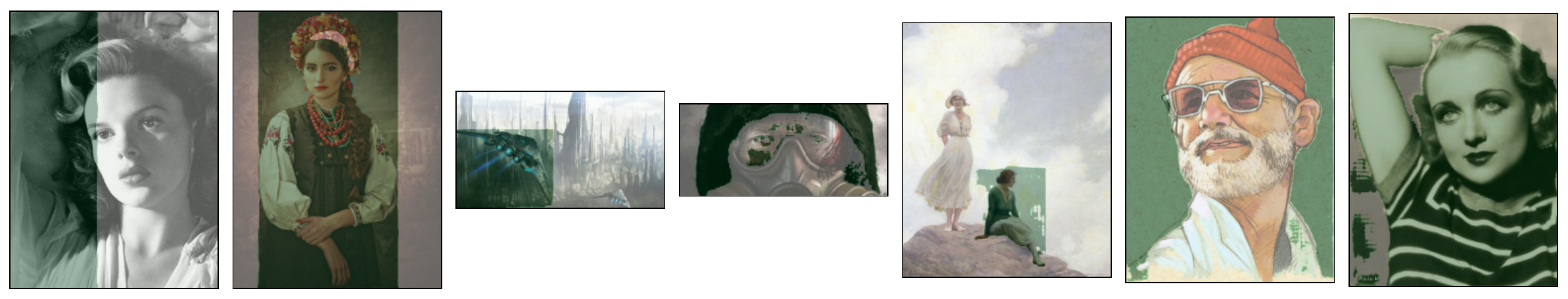}
    \caption{Visualisation of failure modes: segmentation. Green overlay is the estimated instance segmentation.}
    \label{fig:samfail}
\end{figure*}

\begin{figure*}[t]
    \centering
    \includegraphics[width=\linewidth]{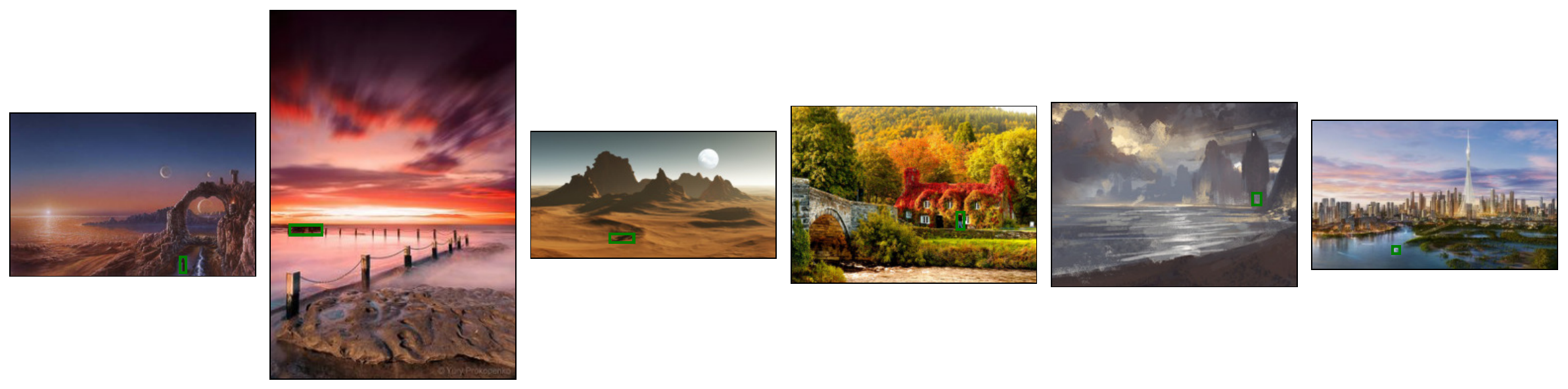}
    \caption{Visualisation of failure modes: irrelevant decomposition. Bounding boxees show detected objects in the image.}
    \label{fig:ptfail}
\end{figure*}

In Figures~\ref{fig:detclipfail}-\ref{fig:ptfail}, we provide several visual examples for all failure modes listed in Sec.~4 of the main paper
: object detection (Fig. \ref{fig:detclipfail}), segmentation (Fig. \ref{fig:samfail}), background (Fig. \ref{fig:bgfail}) and instance inpainting (Fig. \ref{fig:inpaintingfail}), irrelevant decomposition (Fig. \ref{fig:ptfail}). Object detection can comprise missing instances or double detections. The first affects instance and background inpainting, as missed instances are not included in inpainting masks. The second decomposes instances into two or more subcomponents, which often leads to segmentation artefacts. Segmentation limitations can be linked to the SAM model (under or over-segmentation, segmentation of the wrong area within the input bounding box, checkerboard artefacts) or the VIT-Matte model used for alpha layer generation, typically involving over-segmentation. Background issues typically involve introduction of novel instances or structures inconsistent with the overall scene, and text inpainting, despite the use of dedicated negative prompts. This issue often arises when segmentation is imperfect, leaving small artifacts influencing the inpainting process. Similarly, imperfect segmentation and shape priors can influence instance inpainting, leading to failed reconstruction of occluded areas. 

We additionally illustrate limitations of our decomposition pipeline: background occlusions (Fig. \ref{fig:boccfail}) and bounding box constrained inpainting (Fig. \ref{fig:bbfail}). The former occurs because we treat the background as a single flat layer at the bottom of the RGBA stack, while instances can be occluded by background elements (\eg tree branches, large structures). The latter is a limitation of the SAM model, which requires to input local information on where the instance to segment is. We use a dilated version of the object bounding box as input, as leveraging the inpainting mask can lead to segmentation of irrelevant instances. 

\section{Additional Decomposition Results}

\begin{figure*}[t]
    \centering

    \begin{subfigure}[t]{0.875\linewidth}
    \centering\includegraphics[width=\linewidth]{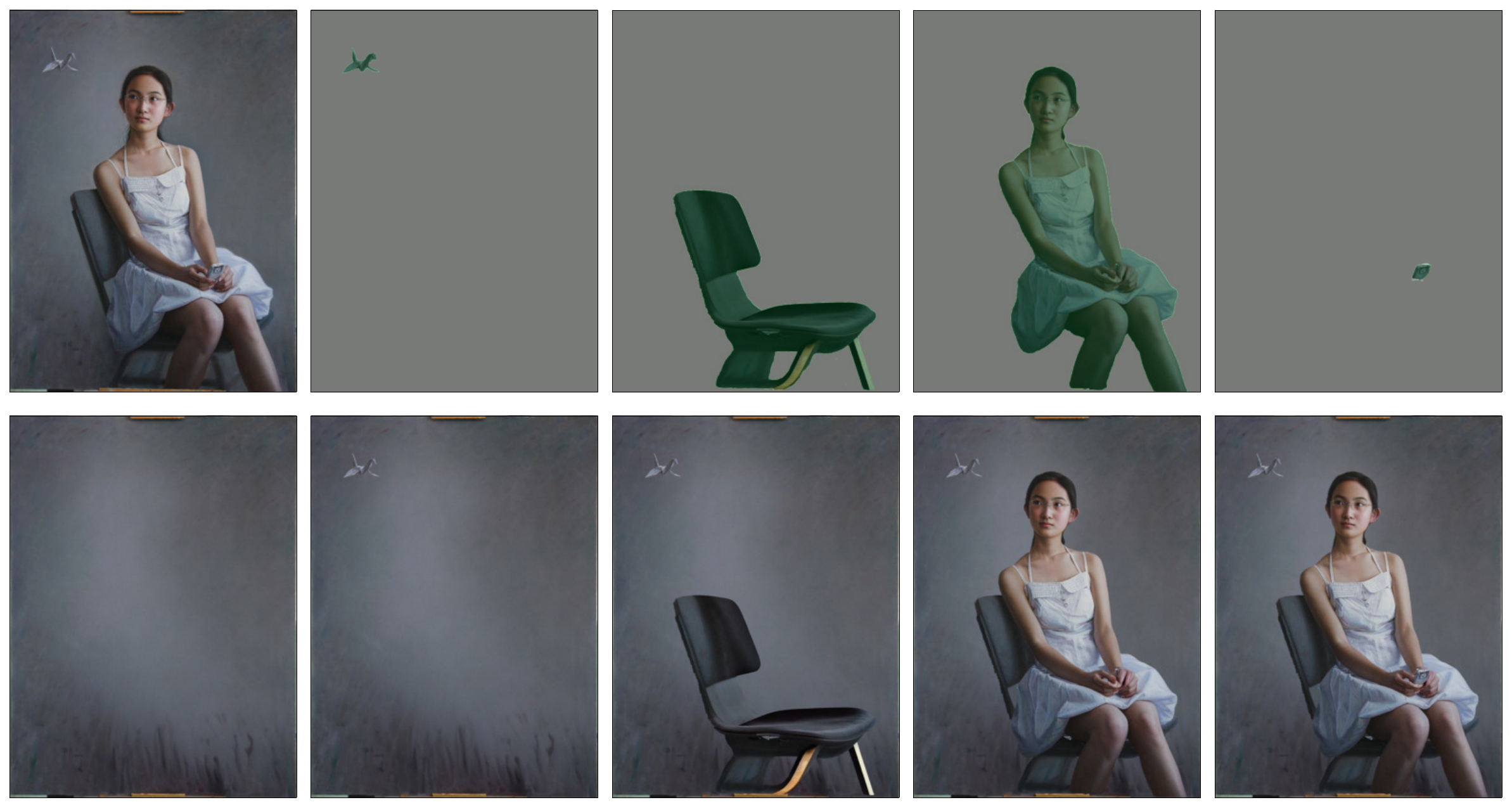}
    \end{subfigure}
    \begin{subfigure}[t]{0.875\linewidth}
    \centering\includegraphics[width=\linewidth]{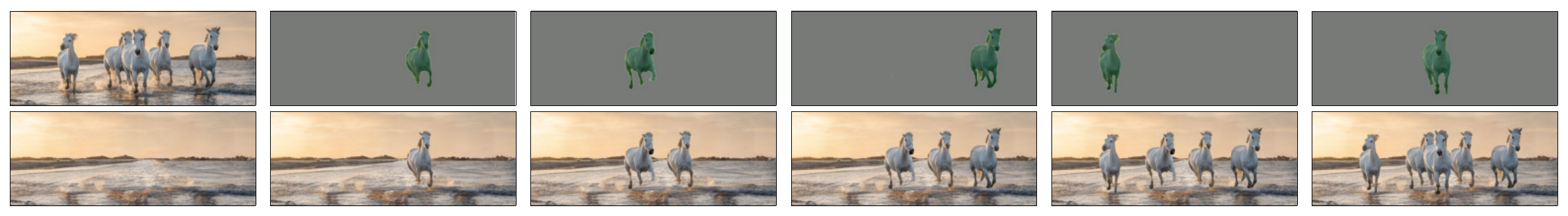}
    \end{subfigure}
    \begin{subfigure}[t]{0.875\linewidth}
    \centering\includegraphics[width=\linewidth]{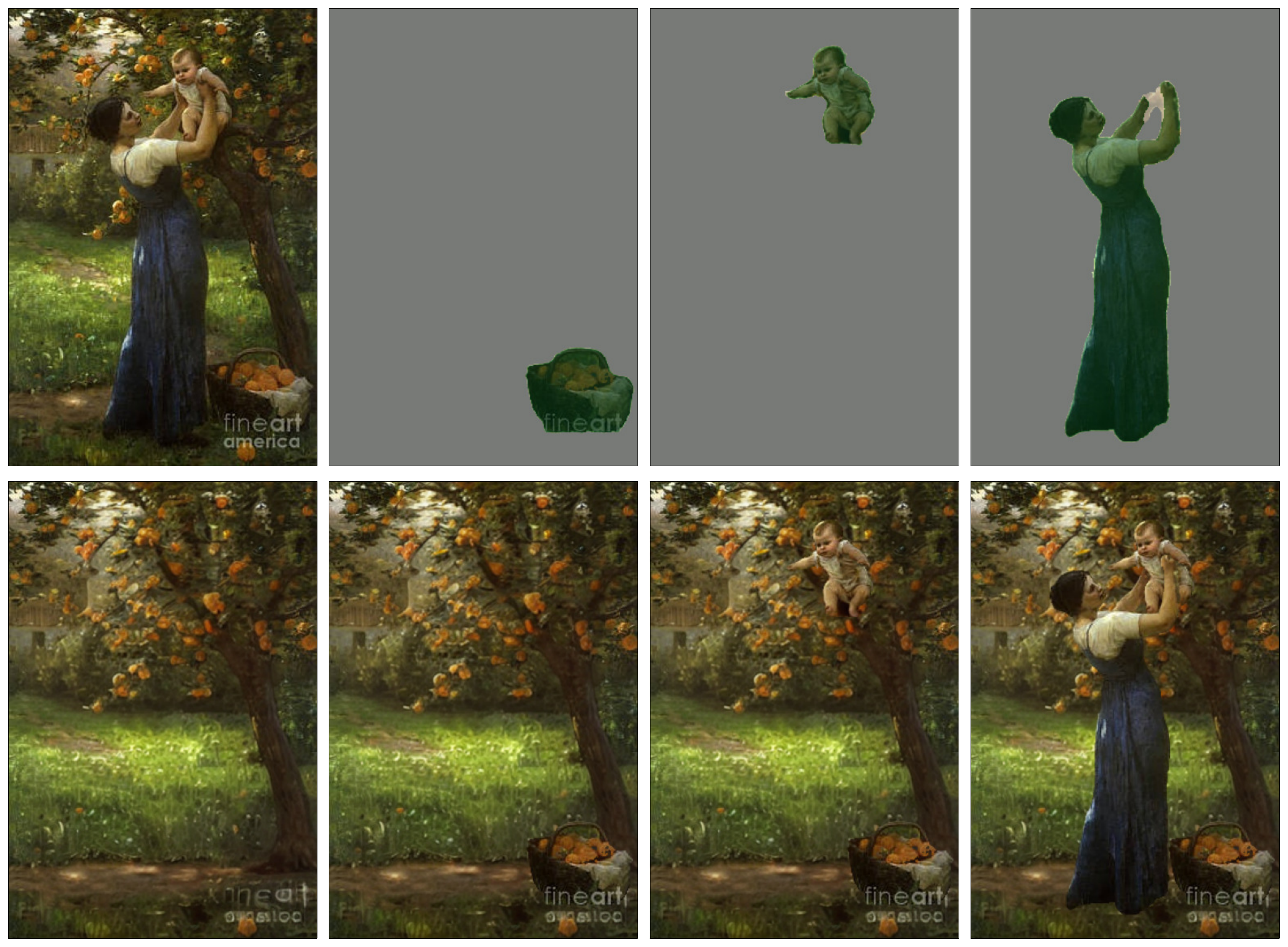}
    \end{subfigure}
    \caption{Visualisation of decomposed images from MuLAn-LAION. For each image, from left to right: original image, instance RGBA image with green alpha overlay (top row); progressively reconstructed image by adding layer one by one (bottom row). }
    \label{fig:resultsLAION}
\end{figure*}

\begin{figure*}[ht!]
    \centering
        \begin{subfigure}[t]{0.9\linewidth}
    \centering\includegraphics[width=\linewidth]{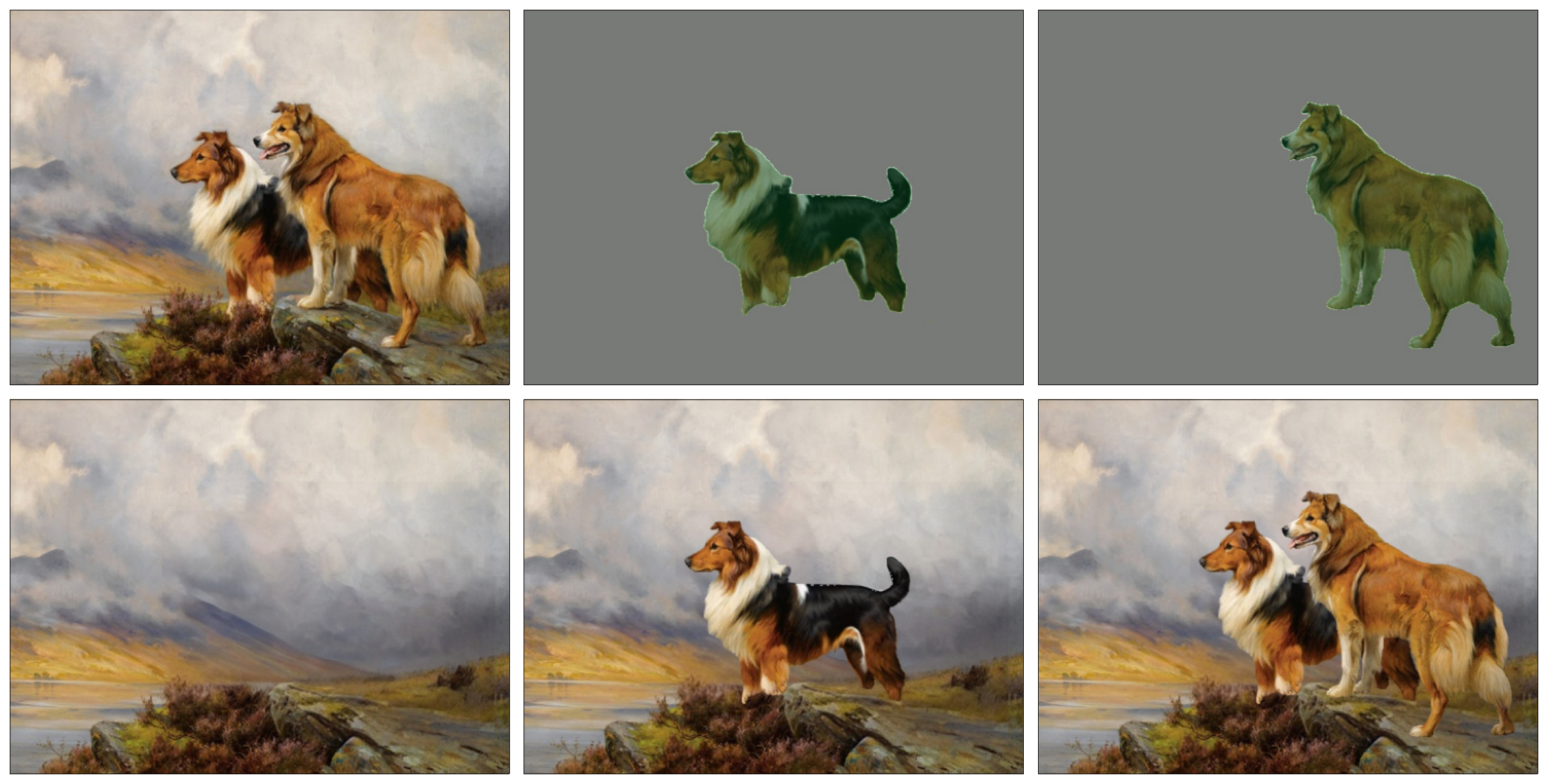}
    \end{subfigure}
        \begin{subfigure}[t]{0.9\linewidth}
    \centering\includegraphics[width=\linewidth]{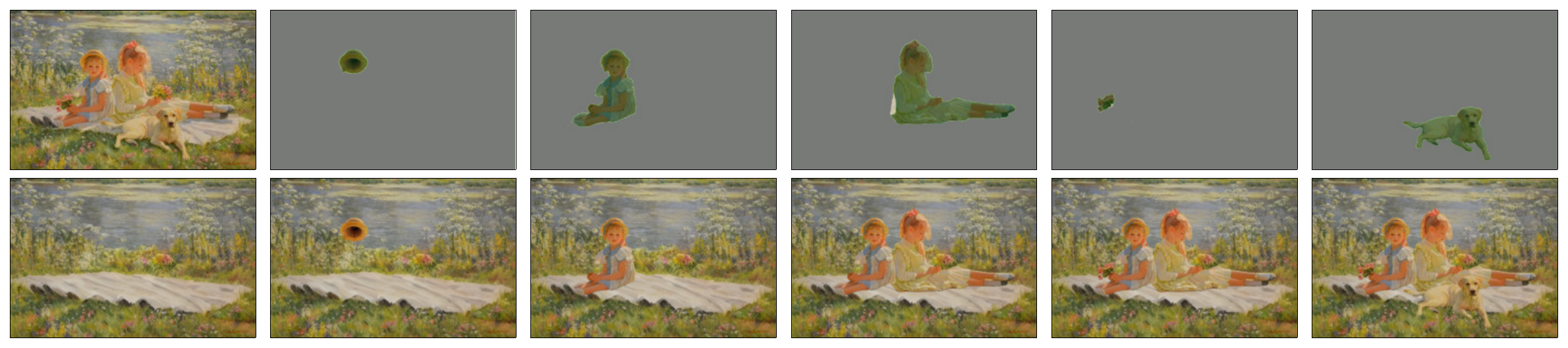}
    \end{subfigure}
        \begin{subfigure}[t]{0.9\linewidth}
    \centering\includegraphics[width=\linewidth]{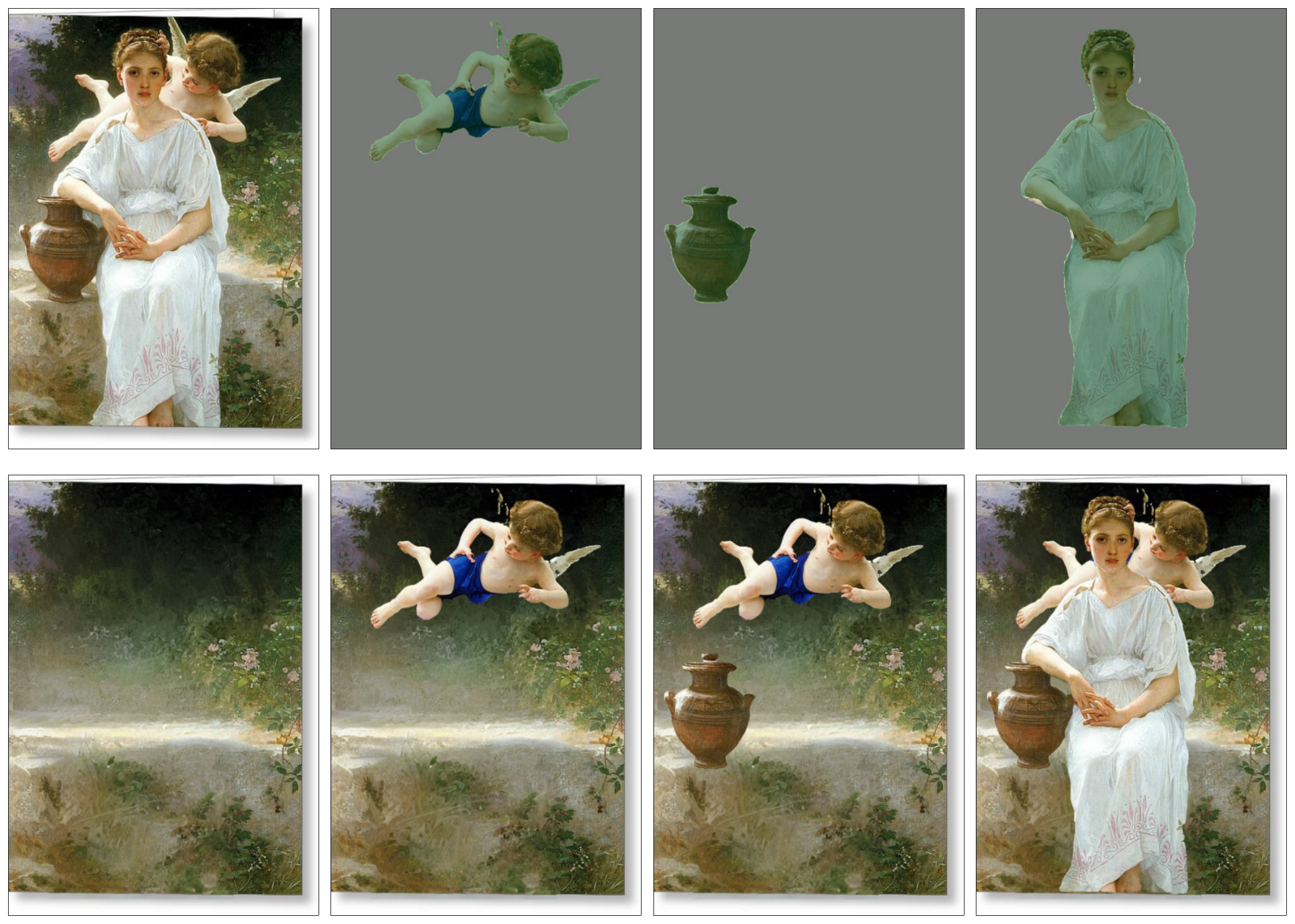}
    \end{subfigure}
    \caption{Visualisation of decomposed images from MuLAn-LAION. For each image, from left to right: original image, instance RGBA image with green alpha overlay (top row); progressively reconstructed image by adding layer one by one (bottom row). }
    \label{fig:resultsLAION2}
\end{figure*}

\begin{figure*}[ht!]
    \centering
    \begin{subfigure}[t]{\linewidth}
    \centering\includegraphics[width=\linewidth]{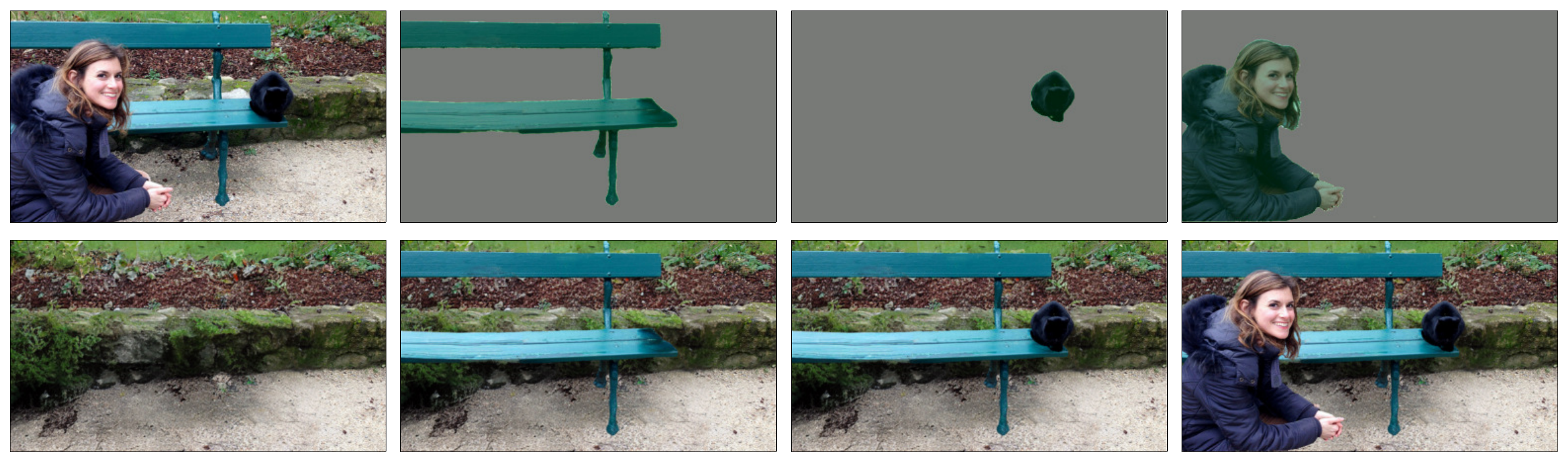}
    \end{subfigure}
        \begin{subfigure}[t]{\linewidth}
    \centering\includegraphics[width=\linewidth]{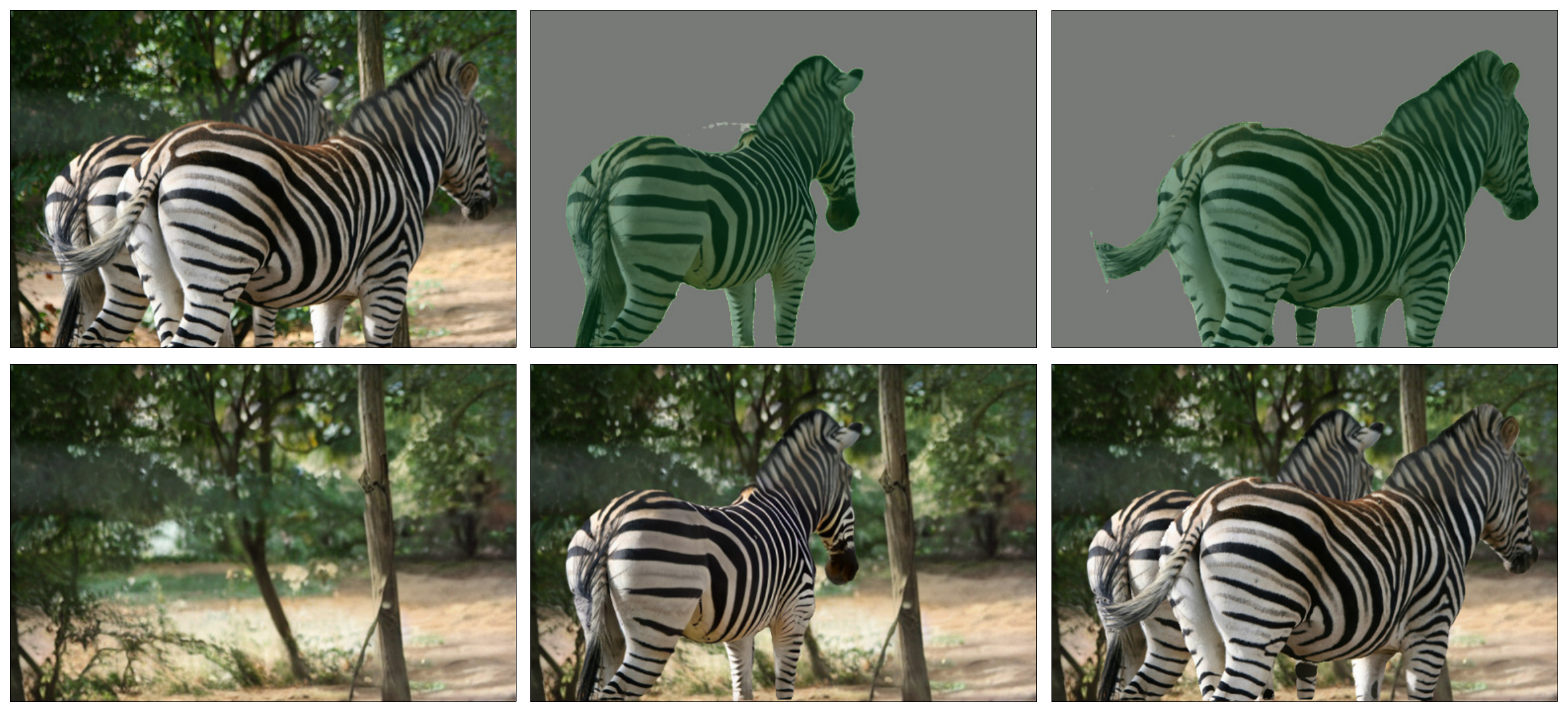}
    \end{subfigure}
    \begin{subfigure}[t]{\linewidth}
    \centering\includegraphics[width=\linewidth]{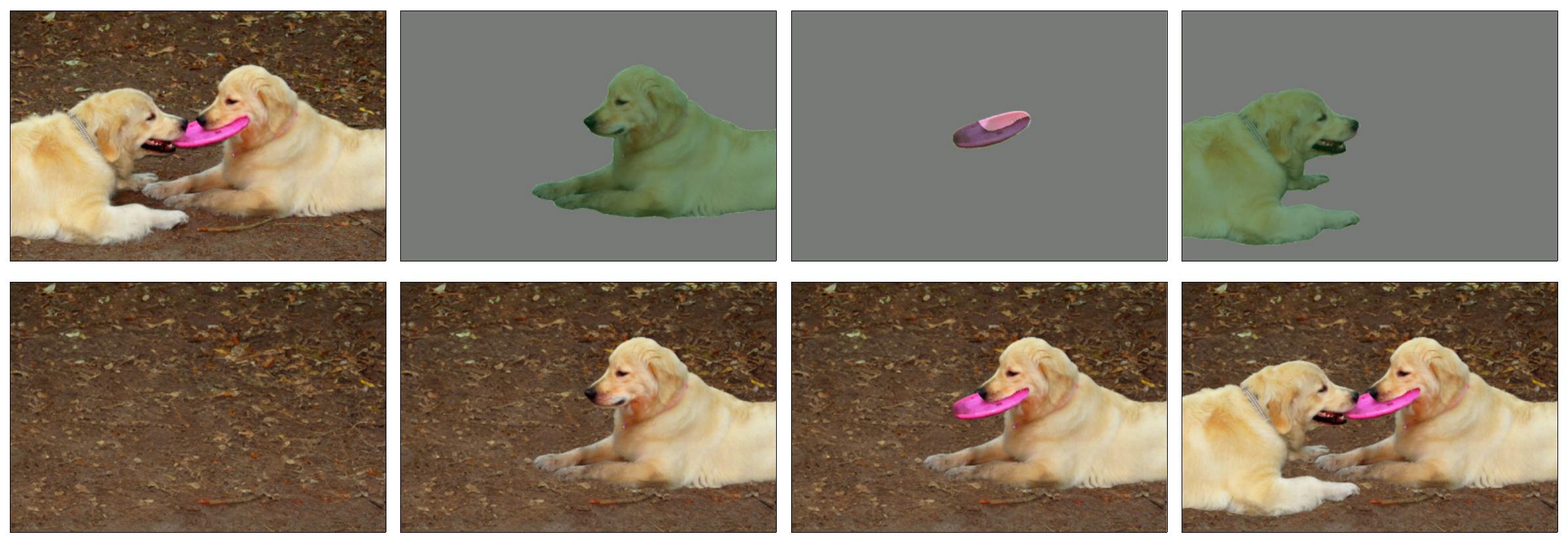}
    \end{subfigure}
        \caption{Visualisation of decomposed images from MuLAn-COCO. For each image, from left to right: original image, instance RGBA image with green alpha overlay (top row); progressively reconstructed image by adding layer one by one (bottom row). }
    \label{fig:resultsCOCO}
\end{figure*}

\begin{figure*}[ht!]
    \centering
    \begin{subfigure}[t]{\linewidth}
    \centering\includegraphics[width=\linewidth]{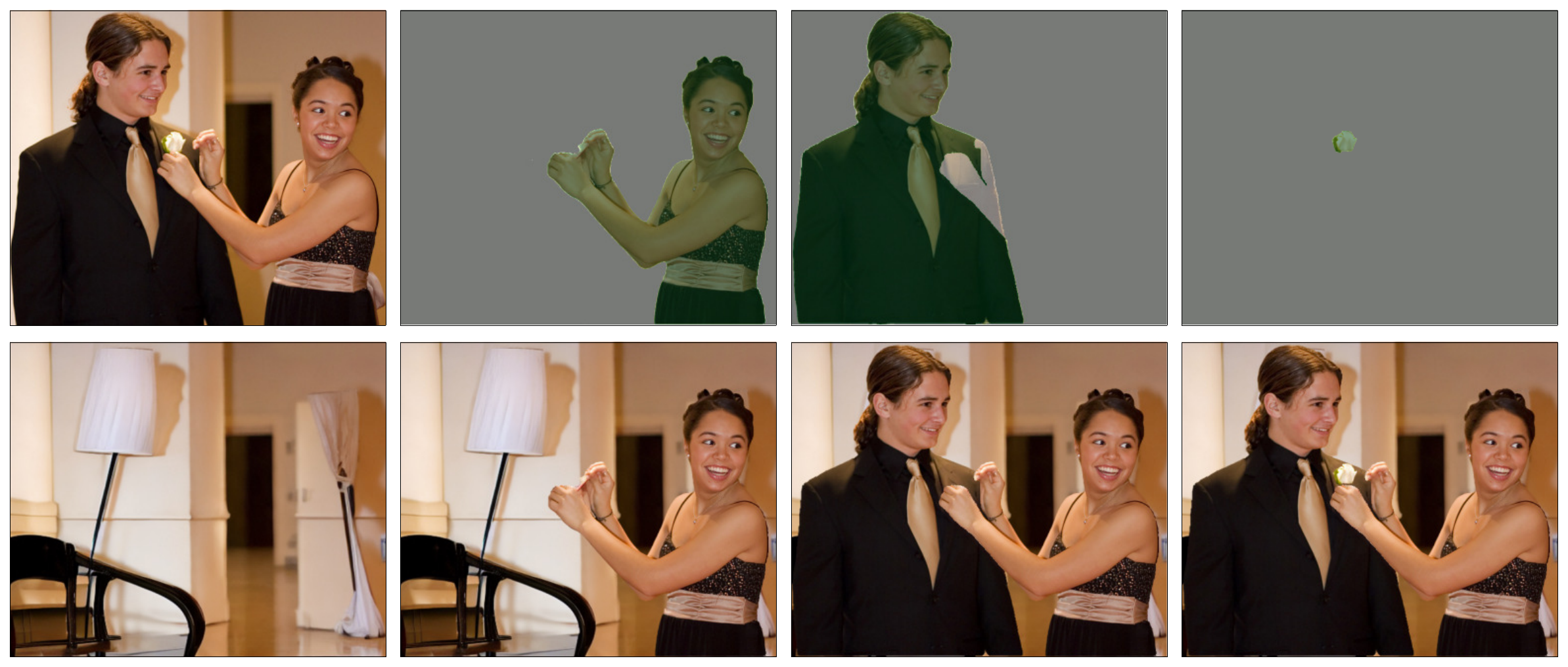}
    \end{subfigure}
    \begin{subfigure}[t]{\linewidth}
    \centering\includegraphics[width=\linewidth]{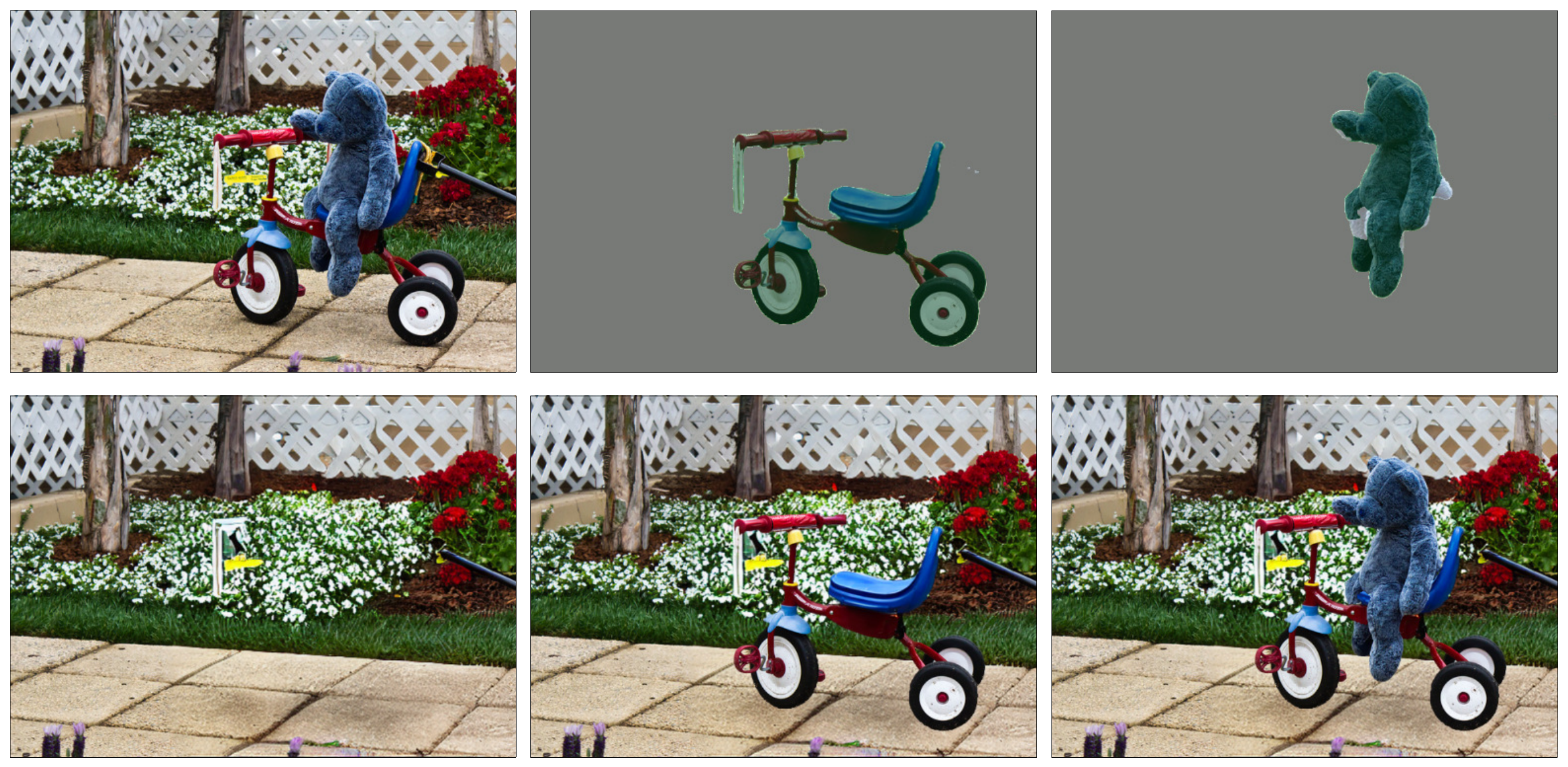}
    \end{subfigure}
        \begin{subfigure}[t]{\linewidth}
    \centering\includegraphics[width=\linewidth]{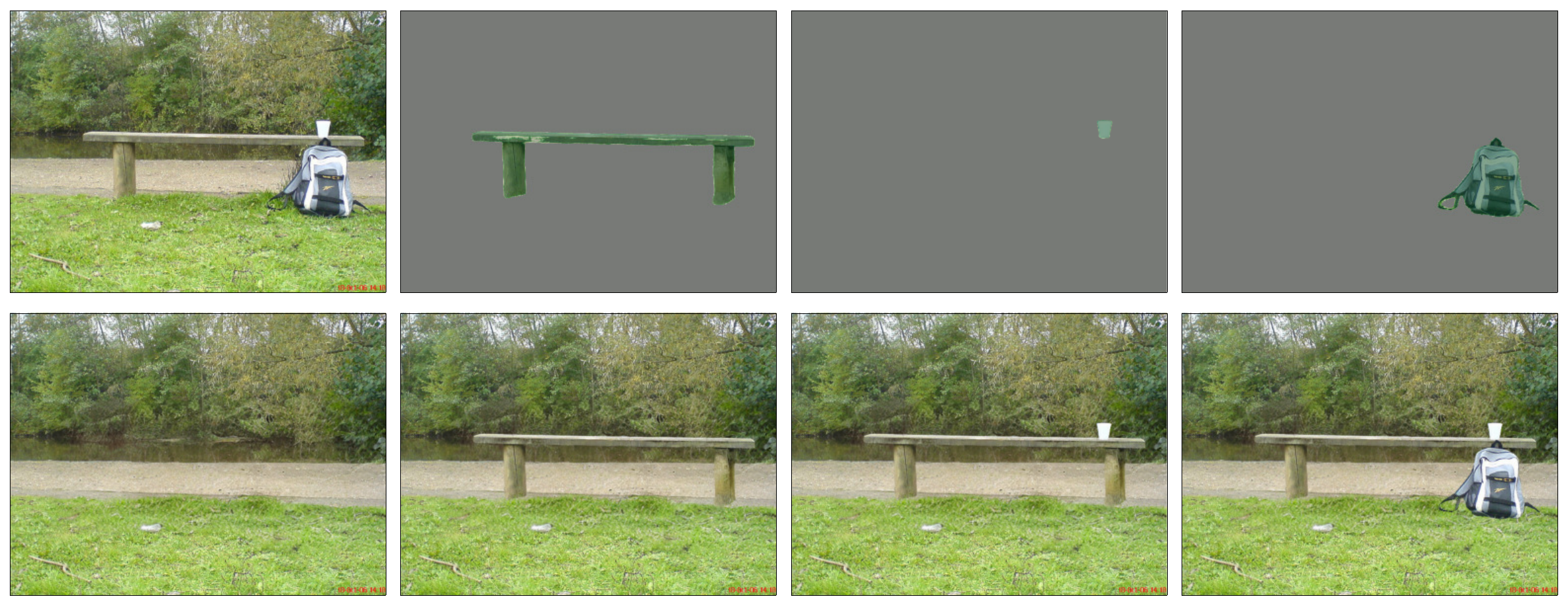}
    \end{subfigure}
    \caption{Visualisation of decomposed images from MuLAn-COCO. For each image, from left to right: original image, instance RGBA image with green alpha overlay (top row); progressively reconstructed image by adding layer one by one (bottom row). }
    \label{fig:resultsCOCO2}
\end{figure*}

Finally, in Figures~\ref{fig:resultsLAION}-\ref{fig:resultsCOCO2}, we show additional visual results of RGBA decompositions in our MuLAn-LAION and MuLAn-COCO datasets. We highlight the varied scenes, image styles and categories. 

\section{Additional Dataset Application Results}

\begin{figure*}[ht!]
    \centering
    \includegraphics[width=\linewidth]{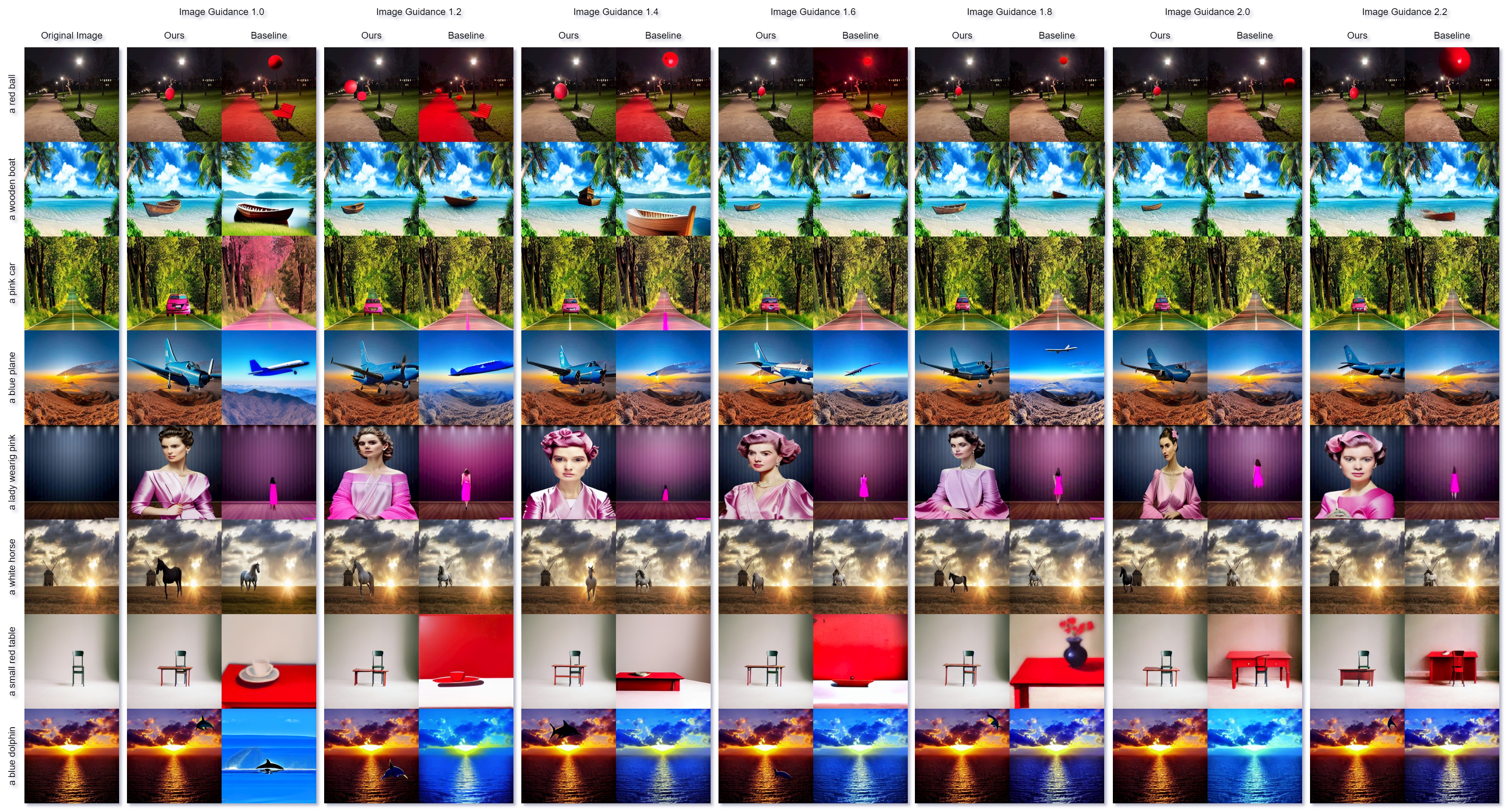}
    \caption{Additional qualitative results of Instance Addition.}
    \label{fig:additional_ip2p}
\end{figure*}

We provide additional visual results for our dataset application experiments. In Fig. \ref{fig:additional_ip2p}, we report qualitative examples of our instance addition experiment, compared to the InstructPix2Pix baseline on our attribute dataset. We can see that we are able to consistently add the desired instance, while at the same time preventing attribute leakage and guaranteeing content preservation.

\begin{figure*}[ht!]
    \centering
    \includegraphics[width=\linewidth]{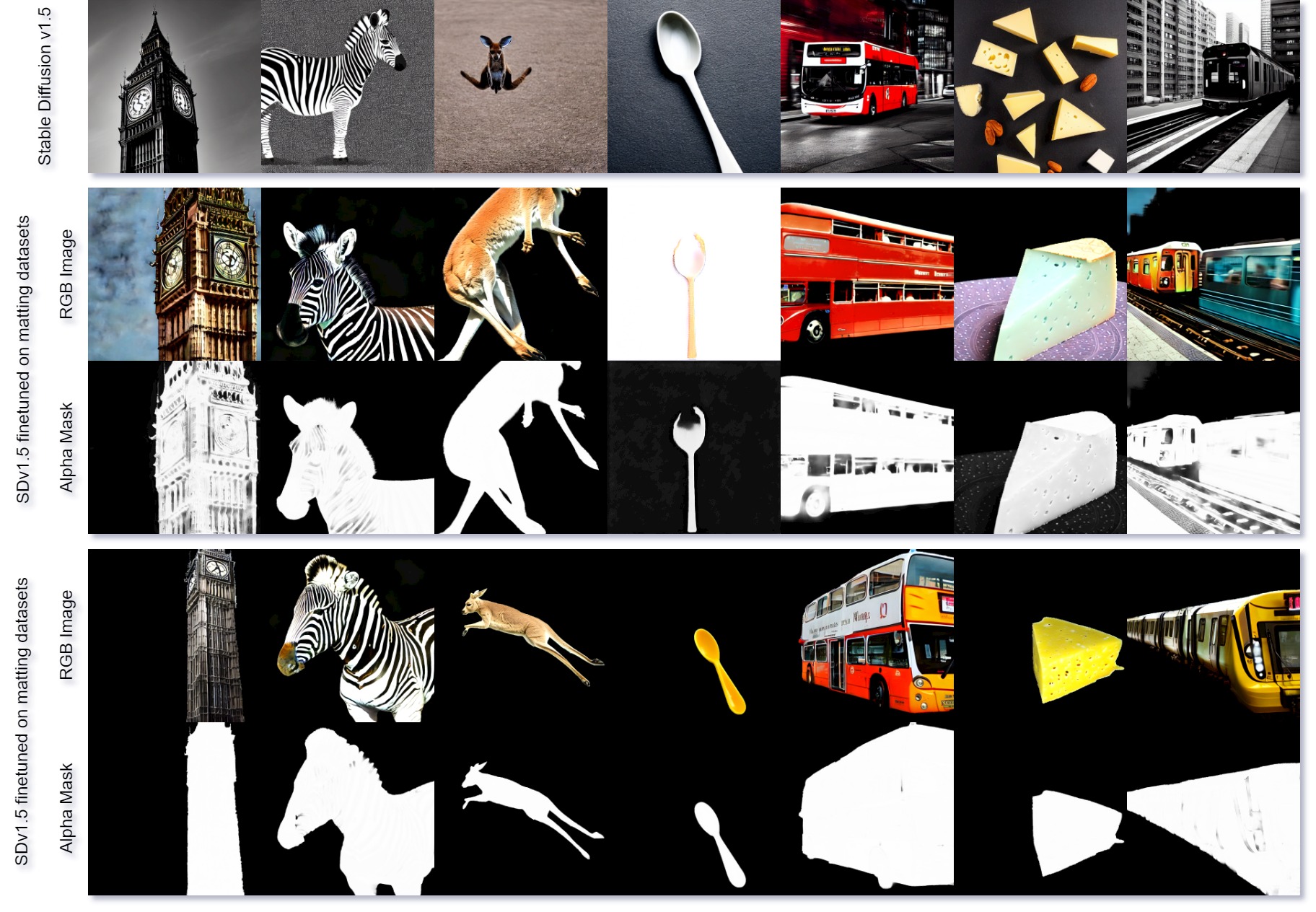}
    \caption{Additional qualitative results of RGBA Generation.}
    \label{fig:additional_rgba}
\end{figure*}

Fig. \ref{fig:additional_rgba} provides additional results for our RGBA generator, compared to our Stable Diffusion baselines (original model and model fine-tuned on matting datasets). Our dataset's diversity allows a better prompt understanding and generation ability, with a better grasp of transparency. Notably, we note that our model trained on matting data tends to include backgrounds in generated instances, and to set black pixels in instances as transparent.

\newpage

\end{document}